# WSense: A Robust Feature Learning Module for Lightweight Human Activity Recognition

Ayokunle Olalekan IGE, Mohd Halim MOHD NOOR

School of Computer Sciences, Universiti Sains Malaysia, 11800 Pulau Pinang, Malaysia
ayokunle.ige@student.usm.my, halimnoor@usm.my

## Abstract

Wearable sensor-based activity recognition involves placing sensors directly on a subject's body to capture positional changes and actions. These wearable sensors are often sampled at a high frequency, at various time steps, and are in time series format. Therefore, there is a need to segment the data into windows before features can be extracted to infer activities. The sliding window approach is the most common for wearable sensor data segmentation. However, the size of sliding window segmentation directly affects the quality of features learned from wearable sensor data and the number of model parameters. In recent times, various modules such as squeeze-and-excitation, and others have been proposed to improve the quality of features learned from wearable sensor signals. However, these modules often cause the number of parameters to be large, which is not suitable for building lightweight human activity recognition models which can be easily deployed on end devices. In this research, we propose a feature learning module, termed WSense, which uses two 1D CNN and global max pooling layers to extract similar quality features from wearable sensor data while ignoring the difference in activity recognition models caused by the size of the sliding window. Experiments were carried out using CNN and ConvLSTM feature learning pipelines on a dataset obtained with a single accelerometer (WISDM) and another obtained using the fusion of accelerometers, gyroscopes, and magnetometers (PAMAP2) under various sliding window sizes. A total of nine hundred sixty (960) experiments were conducted to validate the WSense module against baselines and existing methods on the two datasets. The results showed that the WSense module aided pipelines in learning similar quality features and outperformed the baselines and existing models with a minimal and uniform model size across all sliding window segmentations. The code is available at https://github.com/AOige/WSense.

## 1.0 Introduction

The number of people who are 60 or older is increasing more speedily. According to the global population report, life expectancy at birth is expected to increase to 77 years from 2045–2050 from the current 71 years [1]. Certain approaches have been investigated to preserve or enhance older people's quality of life. Such approaches include creating aged-friendly surroundings and new medical and assistive technology systems to provide long-term care for the elderly [2]. An example of such assistive technology is Human activity recognition (HAR). HAR has been one of the most critical areas of pervasive computing, with applications cutting across various domains such as elder healthcare, childcare, rehabilitation, surveillance, and home automation, among many other areas. HAR aims to understand people's daily behaviour by analyzing the observations collected through a device. Activity recognition can be done through vision-based or sensor-based approaches [3]. The vision-based approach uses images and videos captured during daily human activities by cameras for automatic recognition. However, this method is attributed to privacy issues and coverage distance limitations. Recent advancements in miniaturization [4] have seen the development of low-power, cost-efficient, and high-capacity wireless devices [5]. This has motivated the deployment of wearable devices, which are easier to deploy, cheaper, and can be used over broad areas. Examples of wearable devices include the use of accelerometers, magnetometers, and gyroscopes, among others, for human activity recognition. These wearable devices have led to a whole new approach to activity recognition. They have improved recognition across several domains, such as gait anomaly detection [6], fall detection [7], and animal motion. Various machine-learning models have been proposed to classify human activities obtained from wearable sensors. Trost *et al.* [8], for instance, identified seven activities. Logistic regression was utilized as the classifier by the authors, who used a sensor worn on the wrist and hips to gather activity data. Other machine learning methods that have been explored for the classification of wearable sensor activities include; K-Nearest Neighbour [8], Random Forest [9], Support Vector Machine [10], and Decision Tree [11], among others. However, several limitations of the machine learning approach have led to the adoption of deep learning models capable of extracting features automatically while also improving classification performance.

Generally, the stages of activity recognition in wearable sensors consist of the signal segmentation, feature extraction, and activity classification phases. The signal segmentation phase divides sensor data into smaller chunks called windows, each mapped to a particular activity. This phase is crucial to the performance of activity recognition models because they directly influence the feature learning and classification phases [12,13]. Recently, activity recognition researchers have adopted the sliding window approach to segment wearable sensor datasets. However, the selection of sliding window size often depends on the researcher's expertise and relative hardware constraints. Generally, the window size should be large enough to ensure that at least one activity cycle is contained and that similar activities are distinguishable [14].

Recently, several researchers, as seen in [13,15–17], among others, have investigated the effect of the size sliding window on the classification performance of activity recognition models. It has been established that the size of the sliding window affects feature learning in activity recognition. In general, the window size must catch the relevant signal properties. Short window sizes may divide an activity signal into several distinct windows, which may be considered incomplete. Likewise, larger window sizes may contain activity signals belonging to more than one activity class, which could result in incorrect activity interpretation [12] and also affect feature learning. Since state-of-the-art recognition models rely on precisely segmented data to learn salient features of activity data, it is important to propose methods that can learn quality features from activity data, while ignoring the size of the sliding window. Likewise, the choice of sliding window size directly affects the size of activity recognition models. Large sliding windows often cause activity recognition models to have large parameters.

Due to the limitations of embedded systems, recent activity recognition research has focused on proposing lightweight but efficient activity recognition models that can be easily deployed on end devices [18]. A common approach is the use of attention mechanisms, with the most adopted being the Squeeze and Excitation (SE) block [19]. Several researchers have proposed the SE block and other attention methods to improve the quality of learned features while achieving low model parameters, as seen in [18,20–22], among many others. Even though these attention models improve the quality of learned features, they cannot address the difference in learnt features and the size of activity recognition models with respect to the size of sliding window segmentation. This is because attention mechanisms often calibrate the weight of the extracted features with the weight of the incoming feature maps to extract discriminative features. This is achieved by computing a weighted sum of input features, where the weights are determined dynamically based on some context. This requires learning a set of parameters that can capture the relationships between the input features and the context information, which often comes with increased trainable parameters. However, the literature has not investigated methods to achieve low activity recognition model size while learning similar features regardless of the size of sliding window segmentation.

To contribute, this research introduces a feature learning module termed WSense to learn similar features across all sliding window segmentations from activity signals using lightweight models with uniform sizes. The WSense module uses two 1D convolutional layers and a global max pooling layer to learn similar features from wearable sensor data. The global max pooling layer is applied to the output of the first 1D convolution in the WSense module to obtain a fixed-length feature vector that summarizes the most salient features extracted by feature learning pipelines. The feature maps of the second convolutional layer are then calibrated with the maximum value over each feature map in the first convolutional layer using an element-wise multiplication to increase responsiveness. to ensure improved features are learned. By doing this, fixed features were passed to the fully connected layers of classification models, thereby resulting in a low and uniform number of parameters, regardless of the window segmentation size. Since embedded systems have resource constraints in terms of processor, memory, and battery capacity, keeping the number of model parameters low regardless of the sliding window size will result in easier deployment of activity recognition models on end devices. To the best of our knowledge, our research is the first to exploit ways of improving feature learning while ignoring the difference in learned features and model size caused by choice of sliding window segmentation in human activity recognition. Specifically, our main contributions are in five folds:

i. Firstly, we propose a feature learning module termed WSense to learn better features from wearable sensor data while ignoring the size of sliding window segmentation.
ii. Secondly, the WSense module is integrated to achieve a low and uniform number of parameters across all sliding window sizes.
iii. Thirdly, to demonstrate the extensibility of the WSense module, experiments were conducted on both CNN and ConvLSTM feature learning pipelines.
iv. Extensive experiments were carried out on the baseline models, WSense models, and models integrated with the squeeze and excitation attention, using a dataset obtained with a single accelerometer (WISDM) and another obtained using the fusion of accelerometers, gyroscopes, and magnetometers (PAMAP2).
v. Lastly, the results showed that the WSense module improved the quality of learnt features while learning similar salient features with a low and uniform number of parameters across all sliding window sizes.

The rest of this paper is organized as follows: Section 2 describes the related works, Section 3 presents the research methodology, Section 4 presents the experimental results and discussions, and Section 5 concludes.

## 2.0 Related Works

Researchers have preferred wearable sensors for recognizing human activities because they are cheaper, easier to deploy, and ignore the privacy issues synonymous with other modalities [23]. Research has shown that training HAR models with features extracted from wearable sensors instead of raw signals improve recognition performance [24,25]. As discussed, a significant shortcoming of the machine learning approach is the need to manually extract features from wearable sensor data, as it requires an expert to select features based on heuristics [26]. The limitations attributed to the machine learning approach have seen the adoption of deep learning methods, which can automatically extract features from wearable sensor datasets. Although existing deep learning activity classification models have achieved state-of-the-art performance on benchmarking datasets, the performance on the same benchmarking dataset under various conditions often varies. This variation can be attributed to the choice of data segmentation window and the feature learning approch.

Recently, wearable sensor activity data have been segmented using the activity-defined window method [27] and event-defined window method [28]. The most common and effective approach is the sliding window method. As shown in Figure 1, the sliding window method divides data into fixed-size windows with no gaps between them and, in some cases, overlaps the windows to improve the number of segments [13].

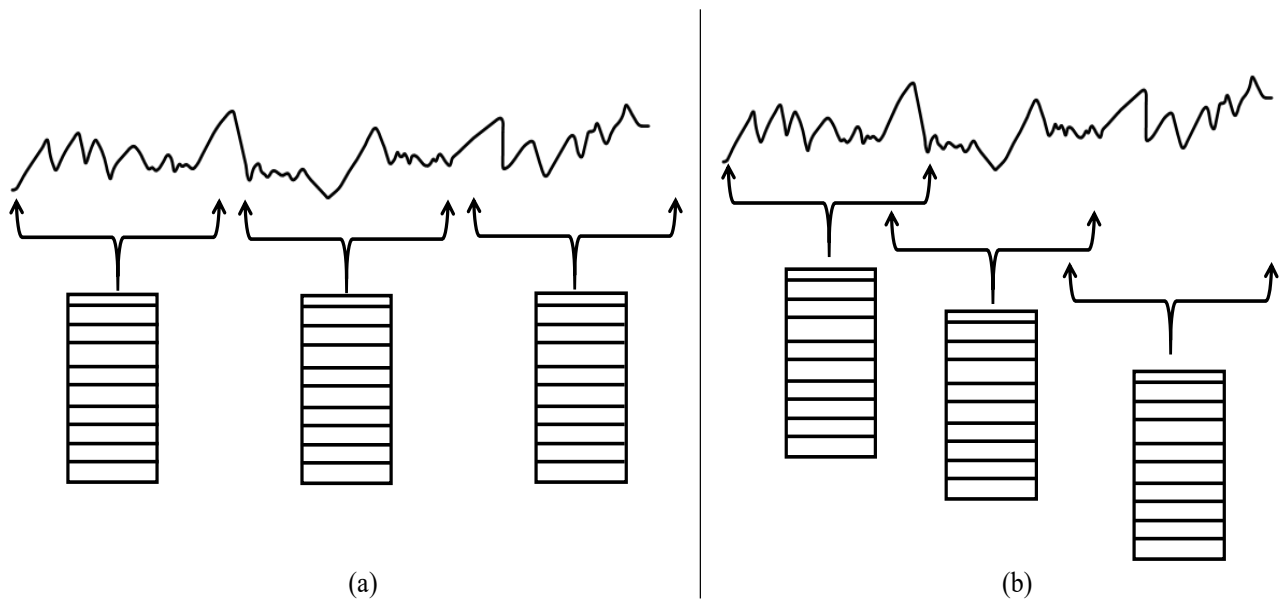

(a)          (b)

Figure 1: (a) Non-Overlapping sliding windows (b) Overlapping sliding windows [23]

In recent times, various window sizes have been adopted by researchers ranging from 0.1 seconds to 13 seconds [12,13,29]. The concept of the sliding window is such that; if sensor data is recorded at a rate of 10 Hz, there will be 10 rows of data for each second activity performed. For a sensor device operating at 10 Hz, we could use a data window representing one second of data, resulting in 10 rows of data. Each activity will have a single window. So, a minute's sensor data will consist of 600 data points or 60 windows of ten-time steps. In Banos *et al.,* [13], extensive experiments were conducted to determine how sliding window size affected activity recognition model performance. The authors concluded that short window sizes achieve good performance between 1 and 2 seconds, while shorter windows have the best performance between 0.25 and 0.5 seconds. Also, in Fida *et al.* [16], the effect of window size was examined regarding the recognition of both short-duration activities, such as sitting and standing, and transitions between activities, and long-duration activities, respectively, respectively, such as walking. According to their research, the optimal recognition trade-off among activities was a 1.5 s window size. A thorough investigation was conducted in [30] to determine how overlapping and non-overlapping sliding windows affect HAR system performance. Based on this, various researchers have employed several window segmentation sizes in wearable sensor-based activity recognition. The following section presents existing works that have adopted various sliding window segmentation approaches in wearable sensor-based human activity recognition.

## 2.1 Activity Recognition Models

In wearable sensor-based activity recognition, signals from accelerometers are often used independently or combined with other sensors such as gyroscopes, magnetometers, and others [23]. Sliding window segmentation of signals obtained using single or multiple accelerometers or a combination of various sensor devices follows the same pattern. Generally, the fixed sliding window with a specific degree of overlap is the most used segmentation method. Several activity recognition models have been developed using this method. For example, in [31], a sliding window of 2.56 seconds with a 50% overlap between two consecutive windows was used for data segmentation on the HAPT dataset [32], which consists of transitional activities. It achieved optimal recognition accuracy of 95.85% using a deep neural network. While on the same dataset, several other researchers have used window sizes ranging from 80 to 180, with 50% overlap. However, some researchers have proposed adaptive sliding windows, which consider the short lengths of transitional activities. An example is in [12] and [33], where the dynamic sliding window approach was used to segment data collected by a single accelerometer. In these works, the window size was changed adaptively to select the most efficient segmentation based on the transition length.

The fixed sliding window with overlap has been the most common approach in datasets without transitions. For example, in [34], the researchers proposed multi-head convolutional attention for activity recognition and tested their model on the WISDM dataset with three window sizes; 48, 64, and 90. Results showed that window size 90 had the highest accuracy at 96.40%. In [35], a capsule neural network was proposed for activity recognition, and a sliding window size of 80 was used on the WISDM dataset. The model achieved an accuracy of 98.67%, but the inclusion of a capsule network in their model suggests a high parameter value since capsule networks have higher model complexity [36]. In [37], a two-stream network is proposed, each representing a spatial and temporal channel. Each stream is integrated with an attention block and a self-attention to deal with multimodal inputs and capture the input sequence's global information. The proposed model is evaluated using four public datasets, among which is PAMAP2, with a window size of 33. The model achieved an accuracy of 98.00%.

In Gao *et al.* [21], a dual attention model was proposed for activity recognition using squeeze and excitation block and experimented on four publicly available datasets, among which are; PAMAP2 and WISDM. The researchers used a sliding window of size 200 on WISDM and 171 on PAMAP2 and achieved an accuracy of 98.85% on WISDM with 2.33M parameters and 93.16% on PAMAP2 with 3.51M parameters. However, similar work in [38] used a sliding window of 128 and achieved a lower performance. Also, in [20], the authors proposed Attsense. They evaluated various sliding window sizes on PAMAP2 and two other datasets, and experiments showed that their model's best result was achieved when a 20-window size was used.

In order to learn quality features from wearable sensor datasets, some researchers have also proposed models based on Recurrent Neural Networks (RNN) and hybrid models, which combine CNN with RNNs such as Gated Recurrent Unit (GRU), Long Short-Term Memory (LSTM), or Bidirectional-LSTM. For example, Dua et al. [39] proposed a multi-input hybrid model that combined CNN with GRU. The model concatenated three CNN-GRU architectures for feature learning and evaluated the model on the PAMAP2, UCI-HAR, and WISDM datasets. A sliding window size of 128 was used to segment the three datasets and achieved 95.27%, 96.20%, and 97.21% accuracies on the three datasets, respectively. Similar work in [40] also proposed a hybrid model to learn features from human activity data using CNN-GRU, evaluated on PAMAP2, WISDM, and UCIHAR, and opted for a sliding window size of 128 on the three datasets. The model achieved recognition accuracy of 96.41% on WISDM, 96.67% on UCI-HAR, and 96.25% on PAMAP2.

In Mutegeki and Han [41], a ConvLSTM approach was used for feature learning. A sliding window size of 128 was used to segment the data, and an accuracy of 91.55% was achieved on the UCIHAR dataset. Khatun et al. [42] proposed a ConvLSTM model with self-attention for feature learning in activity recognition. The model used a sliding window size of 10 to benchmark on two publicly available datasets and one self-collected dataset and achieved recognition accuracy of 98.765 on MHealth and 93.11% on UCIHAR, with 634,052 parameters. In [38], a hybrid model based on CNN with Bi-LSTM was used for feature learning. Experiments on PAMAP2 dataset with 128 sliding window sizes achieved a recognition accuracy of 94.29%. Table 1 presents some existing activity recognition models and their choice of sliding window size on various HAR datasets.

As shown in Table 1 various sliding window sizes have been used on the same HAR datasets on several models, and this has directly affected the performance and size of the model of these models. In this research, we propose the WSense to address the challenges associated with the choice of the sliding window by learning improved and similar features of human activities, regardless of sliding window size, while achieving low and uniform model size across all sliding windows, to allow easy deployment of activity recognition models on end devices.

Table 1: Existing research papers with choice of sliding window size on public HAR datasets

| S/N | Research | Datasets | Window size | Accuracy/F1-score | Parameters |
|---|---|---|---|---|---|
| 1 | Dua et al. [39] | PAMAP2 | 128 | 95.24 | - |
|   |   | WISDM | 128 | 97.21 |   |
| 2 | Xu et al. [43] | PAMAP2 | 171 | 93.50 | - |
| 3 | Gao et al. [21] | WISDM | 200 | 98.85 | 2.33M |
|   |   | UNIMIB-SHAR | 151 | 79.03 | 2.40M |
|   |   | PAMAP2 | 512 | 93.16 | 3.51M |
|   |   | OPPORTUNITY | 64 | 82.75 | 1.57M |
| 4 | Wan et al. [29] | PAMAP2 | 25 | 91.00 | - |
| 5 | Zhang et al. [34] | WISDM | 48, 64, 90 | 95.80, 96.00, 96.40 | - |
| 6 | Nafea et al. [44] | WISDM | 128 | 98.53 | - |
| 7 | Khaled et al. [35] | WISDM | 80 | 98.67% | - |
| 8 | Challa et al. [38] | WISDM | 128 | 96.05 | 0.622M |
|   |   | PAMAP2 | 128 | 94.29 | 0.647M |
| 9 | Ma et al. [20] | PAMAP2 | 20 | 90.60 | - |
| 10 | Han et al. [45] | PAMAP2 | 171 | 92.97 | 1.37M |
|   |   | OPPORTUNITY | 30 | 91.55 | 1.55M |
|   |   | USC-HAD | 512 | 93.49 | 0.42M |
| 11 | Xu et al. [46] | PAMAP2 | 100 | 82.70 | - |
|   |   | WISDM | 200 | 95.78 | - |
| 12 | Xiao et al. [37] | PAMAP2 | 33 | 98.00 | - |

## 3.0 Methodology

In this paper, we propose the WSense module, which can extract salient features from wearable sensor data while ignoring the computational complexity that arises from the choice of sliding window segmentation size. This section presents our methods, and a block diagram of the experimental flow is illustrated in Figure 2.

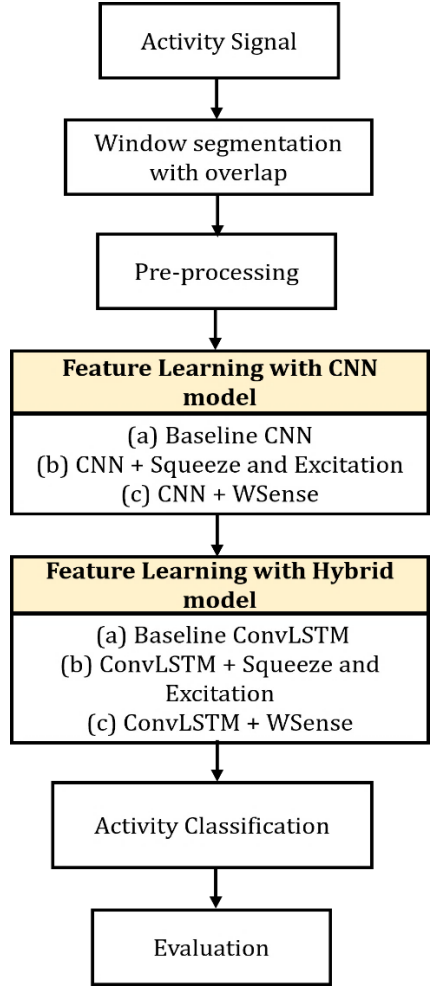

Figure 2: Experimental Flow of the Proposed Method

As shown in Figure 2, the activity signal obtained using wearable sensors is segmented using a sliding window with a degree of overlap, then pre-processed before feeding it to feature learning pipelines. The first feature learning pipeline in the experiment is CNN Feature learning, which used baseline CNN, CNN with Squeeze and Excitation, and CNN with the proposed WSense Module. The second feature learning pipeline in the experiment is the ConvLSTM, which was experimented with using baseline ConvLSTM, ConvLSTM with squeeze and excitation, and ConvLSTM with the proposed WSense. Finally, activity classification is done, and the quality of the features learned on the six pipelines and their model sizes are evaluated based on standard performance metrics. For the Squeeze and excitation model experiment, the neurons of the first and second Fully Connected (FC) layers were set to 128, and an 8 reduction ratio was used in the first FC. The details of the WSense methods are presented in the succeeding sub-sections.

## 3.1 Data Pre-Processing

The activity signals in this research were segmented using a fixed sliding window with a degree of overlap. Given a stream of values $x_i \in \mathbb{S}$ at time $t_i (i \in \mathbb{N})$. It is assumed that $t_0 = 0$, and that the period of sampling is constant at $\Delta T$ [30], such that;

$$\forall i \in \mathbb{N}, t_{i+1} - t_i = \Delta T \quad (1)$$

Using a fixed sliding window size, the signals are split into segments of $n \in \mathbb{N}, n > 1$ sample. Therefore, the window size $T$ can be given as:

$$T = (n-1)\Delta T \quad (2)$$

where $\Delta T$ denotes the sampling period. Given $p \in \{1,2,3 \ldots, n-1\}$ as the number of samples in a certain overlapping period between two consecutive sliding windows, the overlapping period between two consecutive windows in seconds is such that:

$$y = p\Delta T \quad (3)$$

Where the overlapping period is considered as a percentage of the total length of the window and is given as:

$$y(\%) = \frac{p}{n} \quad (4)$$

The overlapping is needed to increase the segmentation numbers to allow better generalization of activity recognition models.

Hence, each sliding window $W_k^c (k \in N)$ can, therefore, be given as a set of values $x_i$, such that:

$$W_k^c = \{x_{k(n-p)}, x_{k(n-p)+1}, x_{k(n-p)+2}, \ldots, x_{k(n-p)+n-1}\}, (k \in \mathbb{N}) \quad (5)$$

where $c$ is the data channels of the sensors.

### 3.2 WSense

The proposed WSense module exploits convolutional and global max pooling layers to capture salient sensor data features while minimizing the number of parameters. A convolutional layer transforms the feature maps $X \in \mathbb{R}^{T' \times C'}$ into the output feature maps $Z \in \mathbb{R}^{T \times C}$ by applying a set of learned filters $F = [f_1, f_2, \ldots, f_C]$. The convolution operation on the input feature maps $X = [x^1, x^2, \ldots, x^{C'}]$ to produce the output feature maps $Z = [z_1, z_2, \ldots z_C]$ is given as:

$$z_i = f_i * X = \sum_{j=1}^{C'} f_i^j * x^j \quad (6)$$

where $f_i = [f_i^1, f_i^2, \ldots, f_i^{C'}]$ and $*$ is the convolutional operation.

The proposed WSense module consists of two 1D convolutional layers with a global max pooling layer in between, followed by reshape and multiplication operations, as shown in Figure 3. The first 1D convolutional layer has a channel dimension $C$ that is equal to the dimension of the incoming channel of the feature learning pipeline. The kernel size is set to 5, and the ELU activation function was used to ensure information under 0 is conserved, and it is given as:

$$f(z) = \begin{cases} z (z \geq 0) \\ \exp(z) - 1 (z < 0) \end{cases} \quad (7)$$

After the first convolutional layer, a global max pooling layer was added to downsample the input representation by taking the maximum value over each feature map in the convolutional layer. By doing this, we were able to consider the maximum output of the features in the convolutional layer and reduce the size of the activations without infringing on the performance of the model.

This automatically allows the features learned in the fully connected layers of the classification model to remain fixed, resulting in a uniform number of parameters, regardless of the window segmentation size. Also, by considering the maximum values of the features maps for all the window sizes, the difference in the learned features is minimized across various sliding window sizes.

The second 1D convolutional layer, which has the same channel dimension as the first, is added to detect the local conjunctions in the preceding feature maps. This layer uses a kernel size of 1 in order to capture more discriminative feature maps since the global max pooling in the preceding layer has summed out the temporal information, and a sigmoid activation function was used in this convolutional layer to retrieve relevant features by ensuring the weight for the feature channel is limited to [0,1]. Then, the maximum value over each feature map in the first convolutional layer are calibrated with the features in the second convolutional layer using an element-wise multiplication to increase responsiveness. The architecture of the WSense is presented in Figure 3.

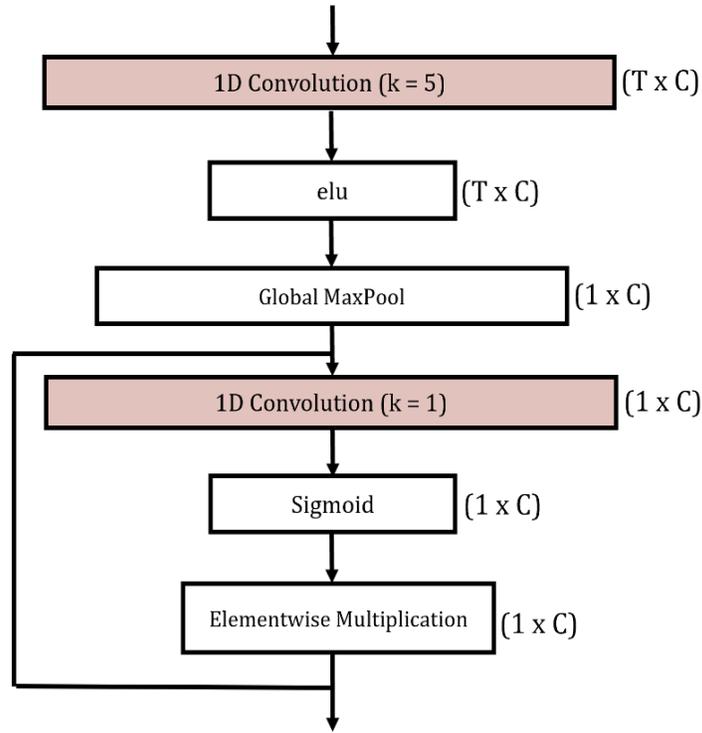

Figure 3: Architecture of the proposed WSense module

## 3.3 Feature Learning Pipelines

### 3.3.1 CNN-WSense

We employed 1D CNN with 32, 64, and 128 channel dimensions in the CNN model, with kernel sizes varying from 3, 5, and 7. The input stage takes in the data input and passes it to the one-dimensional convolutional layer, which uses the ReLU activation function, before a batch normalization layer. A one-dimensional max-pooling layer was then added. The WSense module was placed after the third convolutional layer in the CNN-WSense feature learning pipeline, with 128 channel dimensions. Therefore, the value of the convolutional layers was set to 128 in the WSense module. Then the features were flattened before a dropout layer with a 0.5 dropout rate was included to reduce overfitting. A fully connected layer with 512 neurons and ReLU activation function are placed before the final fully-connected layer with SoftMax activation function to obtain the probability of activity. The probability of activity class $i$ is given as:

$$g(z)_i = \frac{e^{z_i}}{\sum_j^K e^{z_j}} \qquad (8)$$

where $z$ is the input and $K$ is the number of activity classes. The flow diagram of the CNN-WSense classification model is shown in Figure 4.

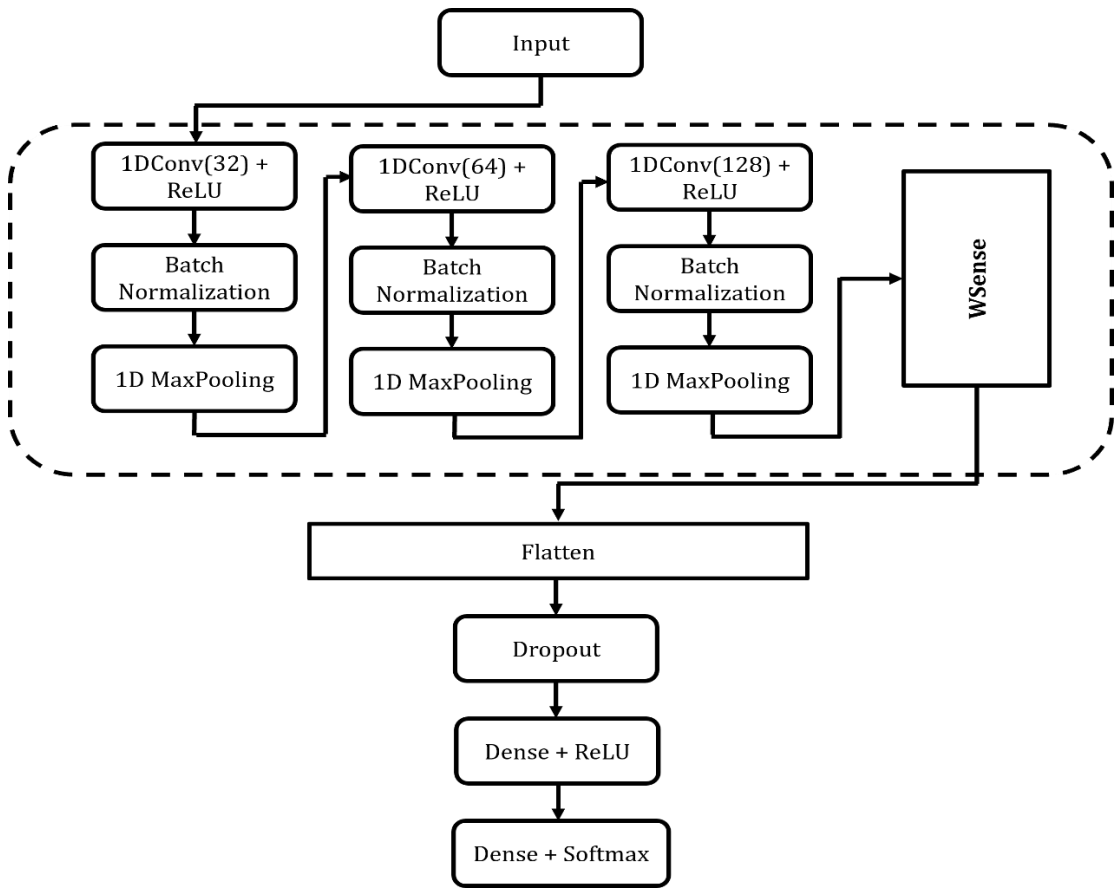

Figure 4: CNN-WSense model

### 3.3.2 ConvLSTM-WSense Model

In the hybrid model, we employed 1D CNN with LSTM layers. The 1D CNN layer has 16, 32, 64, and 128 channel dimensions, with kernel sizes of 1, 3, 5, and 7. Two LSTM layers with tanh activation functions were then added, with 32 and 128 memory cells, to extract the temporal features in the data sequence, and the sequence was returned. The output of the LSTM layer is connected to the WSense module. Since the inclusion of a fully connected layer after the LSTM layer in hybrid feature learning has been shown to minimize training loss, as seen in [41], a fully-connected layer with 512 neurons and ReLU activation function is placed before the final fully-connected layer with SoftMax activation function to obtain the probability of classification. The architecture of the CNN-LSTM-WSense model is presented in Figure 5.

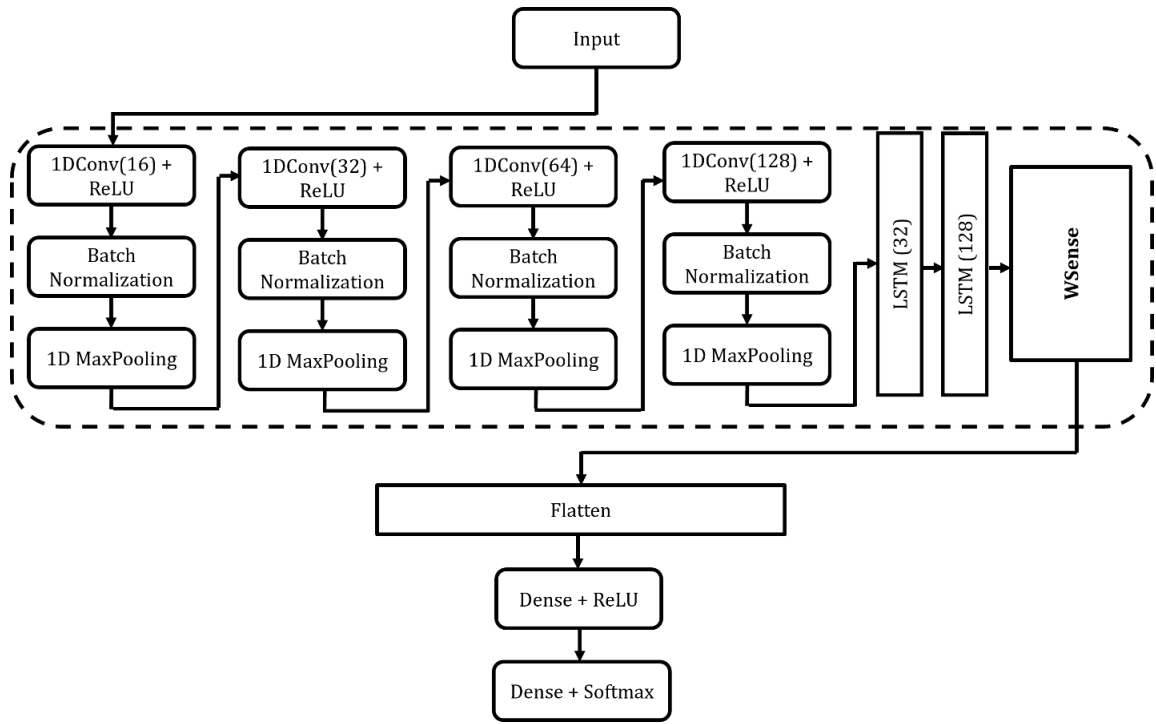

Figure 5: ConvLSTM-WSense

### 4.0 Experimental Results

This section presents the datasets used to evaluate the CNN-WSense and ConvLSTM-WSense models and comparison to the baseline and state-of-the-art models. Also, the implementation details, performance evaluation metrics, and the results on the datasets are presented in this section.

### 4.1 Datasets

#### 4.1.1 WISDM

The WISDM dataset [47] is an activity recognition dataset gathered from 36 participants who go about their daily lives. A single accelerometer data from three-axis was used to obtain this dataset. The dataset consists of 6 activities: Walking, Sitting, Standing, Jogging, Ascending stairs, and Descending stairs. The data was collected at a 20 Hz sampling rate using a smartphone accelerometer sensor.

### 4.1.2 PAMAP2

The PAMAP2 dataset [48] has nine participants, who were all required to participate in twelve protocol activities (including six (6) optional activities). The dataset consists of nine participants' basic and complex activities. Basic activities include sitting, standing, running, descending stairs, ascending stairs, cycling, walking, and Nordic walking, with complex activities such as vacuum cleaning, computer work, car driving, ironing, folding laundry, house cleaning, playing soccer, and rope jumping. Gyroscopes, accelerometers, magnetometers, heart rate monitors, and temperature measurements were used for data collection. In this work, we considered the protocol activities and used 36 features of 3 IMUs.

### 4.2 Implementation Details

The baselines, SE, and models integrated with the proposed WSense module were built using TensorFlow 2.7.0 with Python 3.9 and trained on a workstation equipped with RTX 3050Ti 4GB GPU and 16GB RAM. For a fair comparison, the same workstation, configuration and hyperparameters were used for the Baseline, CNN-SE, ConvLSTM-SE, CNN-WSense and ConvLSTM-WSense models. An epoch of 100 was set, and an early stopping mechanism was used in the callbacks to stop model training once the model stopped improving. During experiments, various sliding window sizes with corresponding overlaps were considered. The choice of sliding window segmentation is based on the literature presented in Section 2. Ten experiments were conducted per sliding window size using the feature learning pipelines plugged with the WSense module, SE block, and corresponding baselines. Taking the total experiments to nine hundred and sixty (960) experiments. After that, the average accuracy of each was calculated, and then the performance was recorded and compared. Each model's 95% Confidence Interval was used to determine the similarity in the quality of the features learnt across the window sizes. The hyperparameters of the models are shown in Table 2.

Table 2: Hyperparameters of the classification models

| Hyperparameters | Details |
| --- | --- |
| Optimizer | Adam |
| Epoch | 100 |
| Batch Size | PAMAP2 – 32, WISDM - 16 |
| Learning rate | Initial Learning rate = $1e^{-4}$ |
| | Minimum Learning rate = $1e^{-7}$ |
| | Patience = 5 |
| Model loss | Categorical cross-entropy |
| | Early stopping patience = 20 |
| Kernel Size | 5, 7, 9 |
| Sliding window size | WISDM – 80, 100, 160, 200, 240, 280, 320, 360 |
| | PAMAP – 171, 250, 300, 360, 400, 450, 500, 550 |
| Sliding window overlap | WISDM – 50% |
| | PAMAP2 – 78% |

## 4.3 Performance Metric

We first considered the number of parameters as an evaluation method, then accuracy, recall, precision, and F1-Score as metrics, and these are given as:

$$Accuracy = \frac{TP + TN}{TP + TN + FP + FN}$$

$$Precision = \frac{TP}{TP + FP}$$

$$Recall = \frac{TP}{TP + FN}$$

$$F1 - Score = \frac{TP}{TP + \frac{1}{2}(FP + FN)}$$

Where TP – True Positive, TN – True Negative, FP – False Positive, and FN – False Negative.

Accuracy is defined as the overall fraction of the activities correctly recognized, Precision is the ratio of positives predicted correctly to the total number of samples classified as positives, Recall is the ratio of accurately predicted positives to the actual number of positive samples, and F1-Score is the harmonic mean of recall and precision. Also, the 95% Confidence Interval is used to measure the difference in the quality of learned features based on the average accuracy achieved across the window sizes.

## 4.4 Experiments on PAMAP2

On the PAMAP2 dataset, ten (10) experiments were conducted on eight window sizes (171, 250, 300, 360, 400, 450, 500, and 550), with the training and testing samples returned per window size presented in Table 3.

Table 3: Training and Testing Sample per selected window size on PAMAP2

|  | Window Size | Training Samples | Testing Samples |
| --- | --- | --- | --- |
|  | 171 | 11445 | 2862 |
|  | 250 | 7776 | 1945 |
|  | 300 | 6460 | 1615 |
| **PAMAP2** | 360 | 5370 | 1343 |
|  | 400 | 4814 | 1204 |
|  | 450 | 4271 | 1068 |
|  | 500 | 3840 | 960 |
|  | 550 | 3476 | 870 |

### 4.4.1 CNN-WSense on PAMAP2

On the CNN-WSense model experiment on the PAMAP2 dataset, experiments were carried out on the baseline model, which excluded the WSense module, on the CNN-SE model, which integrated the Squeeze and excitation block finally, on the CNN-WSense feature learning pipeline. Figure 6 (a) and (b) presents the model training and loss on the best-performing window size.

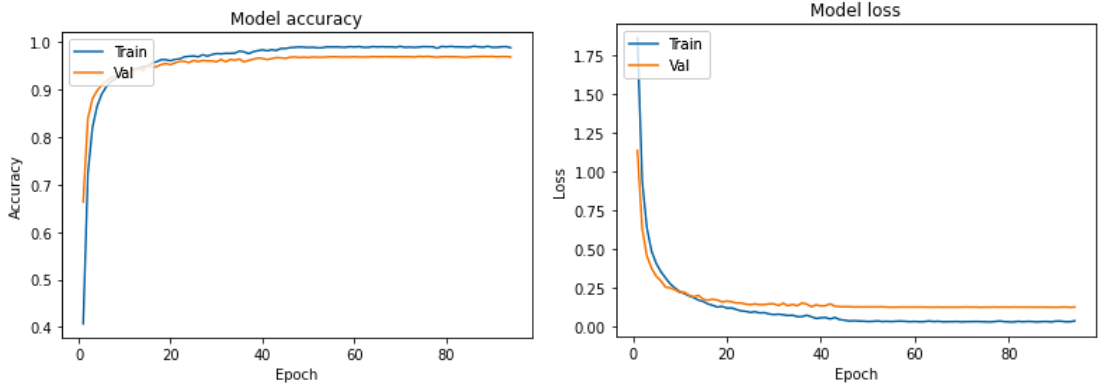

Figure 6: CNN-WSense on PAMAP (a) Model training (b) Model loss

Ten experiments were run on the models for each window size. The average accuracy recorded over the ten experiments per window size and the highest accuracy obtained is shown in Table 4.

Table 4: CNN Comparison of average and highest accuracy on PAMAP2

| Window Size | Baseline CNN | | CNN-SE | | CNN-WSense | |
|---|---|---|---|---|---|---|
| | Average Accuracy(%) | Highest Accuracy(%) | Average Accuracy(%) | Highest Accuracy(%) | Average Accuracy(%) | Highest Accuracy(%) |
| 171 | 96.74 | 96.75 | 97.23 | 97.27 | **97.35** | **97.38** |
| 250 | 96.54 | 96.60 | 96.93 | 96.96 | **97.15** | **97.17** |
| 300 | 95.88 | 95.97 | 96.37 | 96.40 | **97.12** | **97.15** |
| 360 | 95.35 | 95.45 | 96.11 | 96.20 | **96.71** | **96.72** |
| 400 | 96.09 | 96.26 | 95.63 | 95.84 | **96.86** | **96.93** |
| 450 | 93.70 | 93.91 | 95.66 | 95.78 | **97.22** | **97.28** |
| 500 | 92.81 | 93.02 | 94.22 | 94.27 | **96.68** | **96.77** |
| 550 | 93.47 | 93.56 | 96.31 | 96.43 | **97.00** | **97.13** |
| 95% Confidence Interval | 95.07 ± 0.945 | | 96.05 ±0.539 | | **97.01 ±0.141** | |

The average accuracy after ten experiments were carried out per window size presented in Table 4, shows that the baseline CNN model recorded low accuracy compared to the feature learning pipeline of the CNN-WSense. Also, the CNN-SE model was outperformed by the CNN-WSense model on the PAMAP2 dataset. The comparison of the parameters of activity recognition models returned per sliding window size is presented in Table 5.

Table 5: Model size comparison on PAMAP2 (CNN Models)

| | 171 | 250 | 300 | 360 | 400 | 450 | 500 | 550 | Accuracy |
|---|---|---|---|---|---|---|---|---|---|
| CNN | 1.455M | 2.110M | 2.503M | 3.027M | 3.355M | 3.748M | 4.142M | 4.535M | 95.07 ±0.945 |
| CNN-SE | 1.459M | 2.114M | 2.507M | 3.032M | 3.359M | 3.752M | 4.146M | 4.539M | 96.05 ±0.539 |
| **CNN-WSense** | **0.242M** | **0.242M** | **0.242M** | **0.242M** | **0.242M** | **0.242M** | **0.242M** | **0.242M** | **97.01 ±0.141** |

As shown in Table 5, the parameters of the baseline CNN and the CNN-SE models on PAMAP2 increased as the window size increased. The parameters of the baseline CNN ranged between 1.455 million to 4.535 million on window size 171 to 550, which invariably affect the model's training time and deployment on end devices due to the large parameters. Similarly, the CNN-SE model had parameters ranging from 1.459 million on window size 171 to 4.539 million parameters on window size 550. However, the CNN-WSense recorded a minimal parameter value of 0.242 million across all sliding window sizes on the WISDM dataset. This shows that the model size was significantly reduced and remained uniform across all sliding window segmentation sizes. The classification report on experimented windows by the CNN-WSense model is presented in Appendix I.

As shown in Table 4 and Table 5, using the baseline CNN model, window size 171 had a model with 1.455million parameters, and the model achieved average recognition accuracy of 96.74%, while the CNN-SE achieved 97.23% accuracy with 1.459 million parameters. Similarly, window size 250 had a model size of 2.110 million and achieved a recognition accuracy of 96.54%, while the CNN-SE had 2.114 million parameters and recorded an average accuracy of 96.93%. On window size 300, an average recognition accuracy of 95.88% was recorded by the baseline CNN with 2.503 million parameters, while CNN-SE achieved 96.37% with 2.507 million parameters. Likewise, baseline CNN on window size 360 achieved average recognition accuracy of 95.35% with 3.02 million parameters, while CNN-SE had 96.11% with 3.032 million parameters. On window size 400, a model with 3.355 million parameters was returned by the baseline CNN and achieved a recognition accuracy of 96.09%, while CNN-SE had 3.359 million parameters with 95.63% average recognition accuracy.

Also, window size 450 had an average recognition accuracy of 93.70% with 3.748 million parameters using the baseline CNN, while the CNN-SE had 95.66% average accuracy with 3.752 million parameters. Likewise, on 500 sliding window size, an average recognition accuracy of 92.81% was recorded by the baseline CNN, using a model with 4.142 million parameters, while the CNN-SE model had 4.146 million parameters and recorded 94.22% average accuracy. Lastly, on sliding window size 550, the baseline CNN recorded an average recognition accuracy of 93.47%, with 4.535 million parameters, while CNN-SE, with 4.539 million parameters, had an average recognition accuracy of 96.31%. However, by using the CNN-WSense feature learning pipeline on PAMAP2, the average and highest recognition accuracy outperformed both the baseline CNN and CNN-SE models across all experimented sliding window sizes, and a minimal and uniform model size of 242,924 (0.242million) was recorded across all the sliding window sizes, as shown in Table 5. The average accuracy comparison is presented in Figure 7.

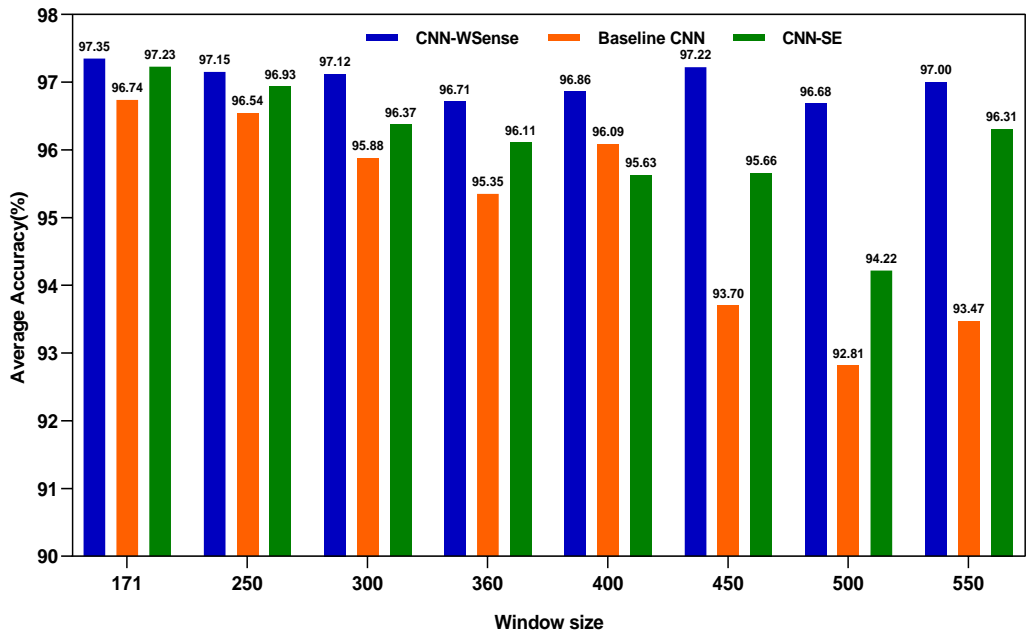

Figure 7: Average accuracy comparison of CNN models on PAMAP2 window sizes

The confusion matrix of the highest recognition accuracy achieved by the CNN-WSense model on PAMAP2 is presented in Figure 8.

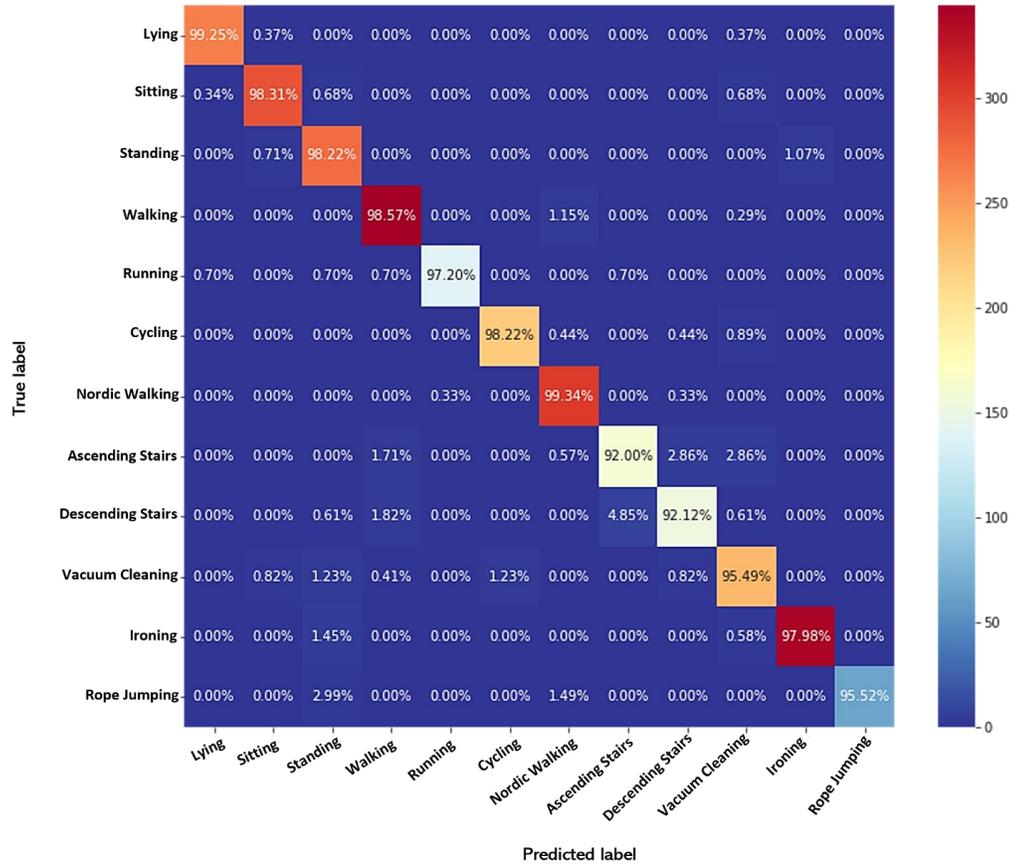

Figure 8: Confusion Matrix of 171 window size (CNN-WSense on PAMAP2)

As shown in the confusion matrix shown in Figure 8, lying activity had 99.25% of its samples correctly classified, with 0.37% of lying samples misclassified as sitting and vacuum cleaning. On sitting activity, 98.31% of the samples were classified correctly, with 0.34% misclassified as lying, 0.68% as standing, and another 0.68% misclassified as vacuum cleaning. On standing activity, 98.22% of its samples were correctly classified, with 0.71% misclassified as sitting and 1.07% misclassified as ironing. Similarly, walking activity had 98.57% of samples classified correctly, with 1.15% misclassified as Nordic walking and 0.29% misclassified as vacuum cleaning. Running activity had 97.20% of its samples classified correctly, with 0.70% misclassified as lying, standing, walking, and descending stairs.

Cycling had 98.22% of samples classified correctly, with 0.44% misclassified as Nordic walking and downstairs, while 0.89% was misclassified as vacuum cleaning. Ascending stairs had 92.00% of its samples correctly classified, with 1.71% misclassified as walking, 0.57% misclassified as Nordic walking, 2.86% as descending stairs, and another 2.86% of the samples misclassified as vacuum cleaning. Similar to the result on ascending stairs, descending stairs activity had 92.12% of its samples correctly classified, with 1.82% misclassified as walking, 4.85% misclassified as ascending stairs, and 0.61% misclassified as vacuum cleaning and standing. On the Rope jumping activity, 95.52% of the samples were correctly classified, with 2.99% misclassified as standing and 1.49% as Nordic walking.

### 4.4.2 ConvLSTM-WSense on PAMAP2

Also, on the ConvLSTM-WSense model experiment on the PAMAP2 dataset, experiments were carried out on the baseline model, which excluded the WSense module. This CNN-SE model integrated the squeeze and excitation block and, finally, on the CNN-WSense feature learning pipeline. Figures 9 (a) and (b) present the model training and loss on the best-performing window size.

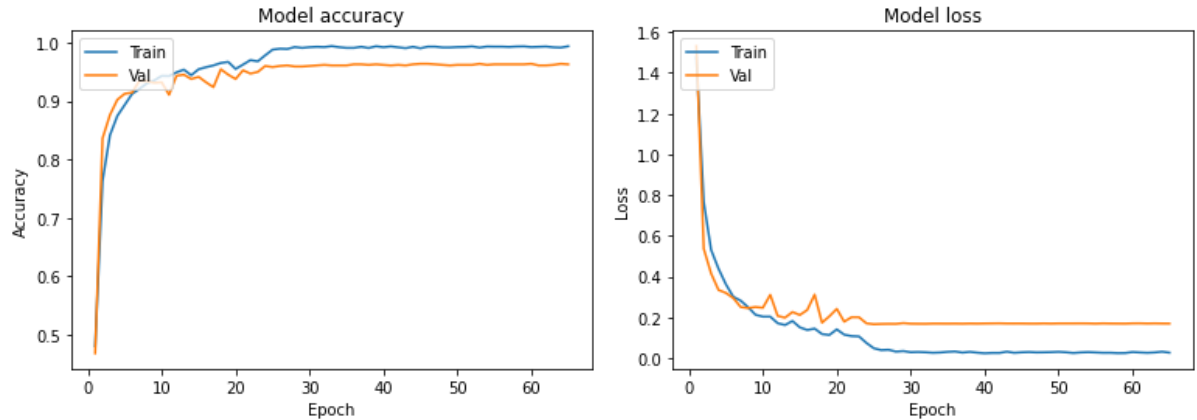

Figure 9: ConvLSTM-WSense on PAMAP2 (a) Model training (b) Model loss

Ten experiments were run on the models for each window size. The average accuracy recorded over the ten experiments per window size and the highest accuracy obtained is shown in Table 6.

Table 6: ConvLSTM Comparison of average and highest accuracy on PAMAP2

|  | Baseline ConvLSTM | | ConvLSTM-SE | | ConvLSTM-WSense | |
| --- | --- | --- | --- | --- | --- | --- |
| Window Size | Average Accuracy(%) | Highest Accuracy(%) | Average Accuracy(%) | Highest Accuracy(%) | Average Accuracy(%) | Highest Accuracy(%) |
| 171 | 96.45 | 96.47 | 96.86 | 96.89 | **97.39** | 97.41 |
| 250 | 96.53 | 96.55 | 96.71 | 96.76 | **96.83** | 96.86 |
| 300 | 95.92 | 95.97 | 96.26 | 96.28 | **96.35** | 96.41 |
| 360 | 95.02 | 95.09 | 96.19 | 96.20 | **96.45** | 96.45 |
| 400 | 96.33 | 96.34 | 96.68 | 96.76 | **96.78** | 96.84 |
| 450 | 95.37 | 95.50 | 96.15 | 96.25 | **96.99** | 97.00 |
| 500 | 95.20 | 95.20 | 95.53 | 95.62 | **96.40** | 96.56 |
| 550 | 95.52 | 95.63 | 95.97 | 95.97 | **96.78** | 96.78 |
| 95% Confidence Interval | 95.79 ± 0.413 | | 96.29 ±0.257 | | **96.75 ± 0.241** | |

The average accuracy after ten experiments were carried out per window size presented in Table 6 shows that the baseline ConvLSTM model recorded low accuracy compared to the ConvLSTM-WSense model. Also, the ConvLSTM-SE model outperformed the ConvLSTM-WSense on the PAMAP2 dataset. The comparison of the parameters of activity recognition models returned per sliding window size on PAMAP2 is presented in Table 7.

Table 7: Model Comparison across window sizes (ConvLSTM models on PAMAP2)

|  | 171 | 250 | 300 | 360 | 400 | 450 | 500 | 550 | Accuracy |
|---|---|---|---|---|---|---|---|---|---|
| ConvLSTM | 0.835M | 1.163M | 1.360M | 1.622M | 1.819M | 2.015M | 2.212M | 2.408M | 95.79 ±0.413 |
| ConvLSTM-SE | 0.840M | 1.167M | 1.364M | 1.626M | 1.823M | 2.019M | 2.216M | 2.412M | 96.29 ±0.257 |
| **ConvLSTM-WSense** | **0.344M** | **0.344M** | **0.344M** | **0.344M** | **0.344M** | **0.344M** | **0.344M** | **0.344M** | **96.75 ±0.241** |

As shown in Table 7, the parameters of the baseline ConvLSTM and the ConvLSTM-SE models on PAMAP2 increased as the window size increased. However, the CNN-WSense recorded a minimal parameter value of 0.344 million across all sliding window sizes. This shows that the model size was significantly reduced and remained uniform across all sliding window segmentation sizes. The classification report on experimented windows is presented in Figure 11.

As shown in Table 6 and Table 7, using the baseline ConvLSTM model, window size 171 had a model with 0.835million parameters, and the model achieved an average recognition accuracy of 96.45%, with the highest being 96.47%, while the ConvLSTM-SE achieved 96.89% accuracy with 0.840million parameters, with the highest accuracy being 96.89. Window size 250 had a model size of 1.163 million on baseline ConvLSTM. It achieved a recognition accuracy of 96.53%, with the highest accuracy being 96.55%, while the CNN-SE had 1.167 million parameters and recorded an average accuracy of 96.71%. On window size 300, an average recognition accuracy of 95.92% was recorded by the baseline ConvLSTM with 1.360 million parameters, while ConvLSTM-SE achieved an average recognition accuracy of 96.26% with 1.364 million parameters. Likewise, baseline ConvLSTM on window size 360 achieved average recognition accuracy of 95.02% with 1.622 million parameters, while ConvLSTM-SE had 96.19% with 1.626 million parameters. Experiments on window size 400 returned a model with 1.819 million parameters on the baseline ConvLSTM and achieved a recognition accuracy of 96.33%, while ConvLSTM-SE had 1.823 million parameters with 96.68% average recognition accuracy.

Also, window size 450 had an average recognition accuracy of 95.37% with 2.015 million parameters using the baseline ConvLSTM, while the ConvLSTM-SE had 96.15% average accuracy with 2.019 million parameters. On 500 sliding window size, an average recognition accuracy of 95.20% was recorded by the baseline ConvLSTM, using a model with 2.212million parameters, while the ConvLSTM-SE model had 2.216million parameters and recorded 95.53% average recognition accuracy. Lastly, on sliding window size 550, an average recognition accuracy of 95.52% was recorded by the baseline ConvLSTM, with 2.408million parameters, while ConvLSTM-SE, with 2.412 million parameters, had an average recognition accuracy of 95.97%. However, by using the ConvLSTM-WSense model on PAMAP2, the average and highest recognition accuracy outperformed both the baseline ConvLSTM and ConvLSTM-SE models across all experimented sliding window size, and a minimal and uniform model size of 344,700 (0.344million) was recorded across all the sliding window sizes, as shown in Table 10. The average accuracy comparison is presented in Figure 10.

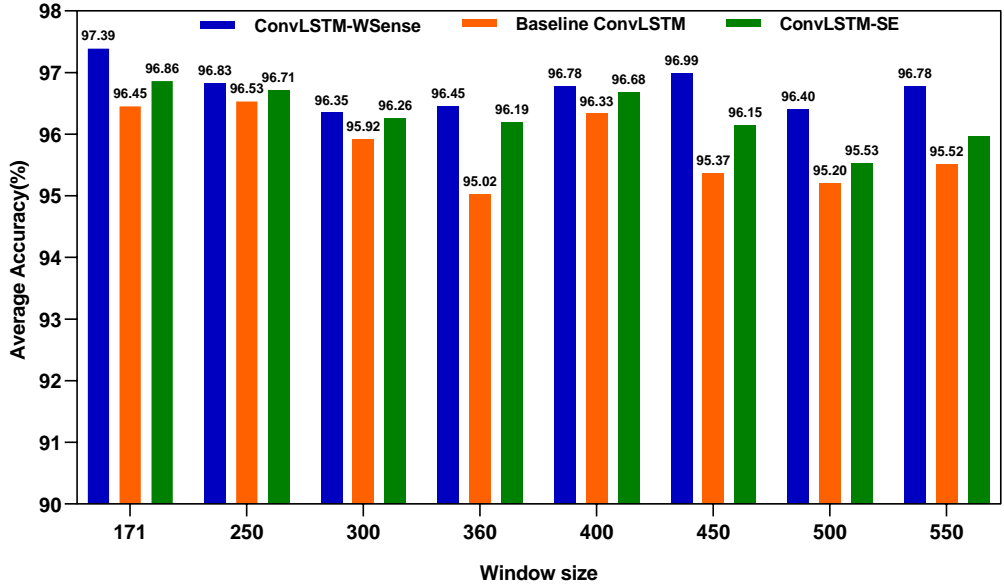

Figure 10: Comparison of average accuracy per window size of ConvLSTM models on PAMAP2

The confusion matrix of the highest recognition accuracy recorded across the eight experimented window sizes on the ConvLSTM-WSense feature learning pipeline is shown in Figure 11.

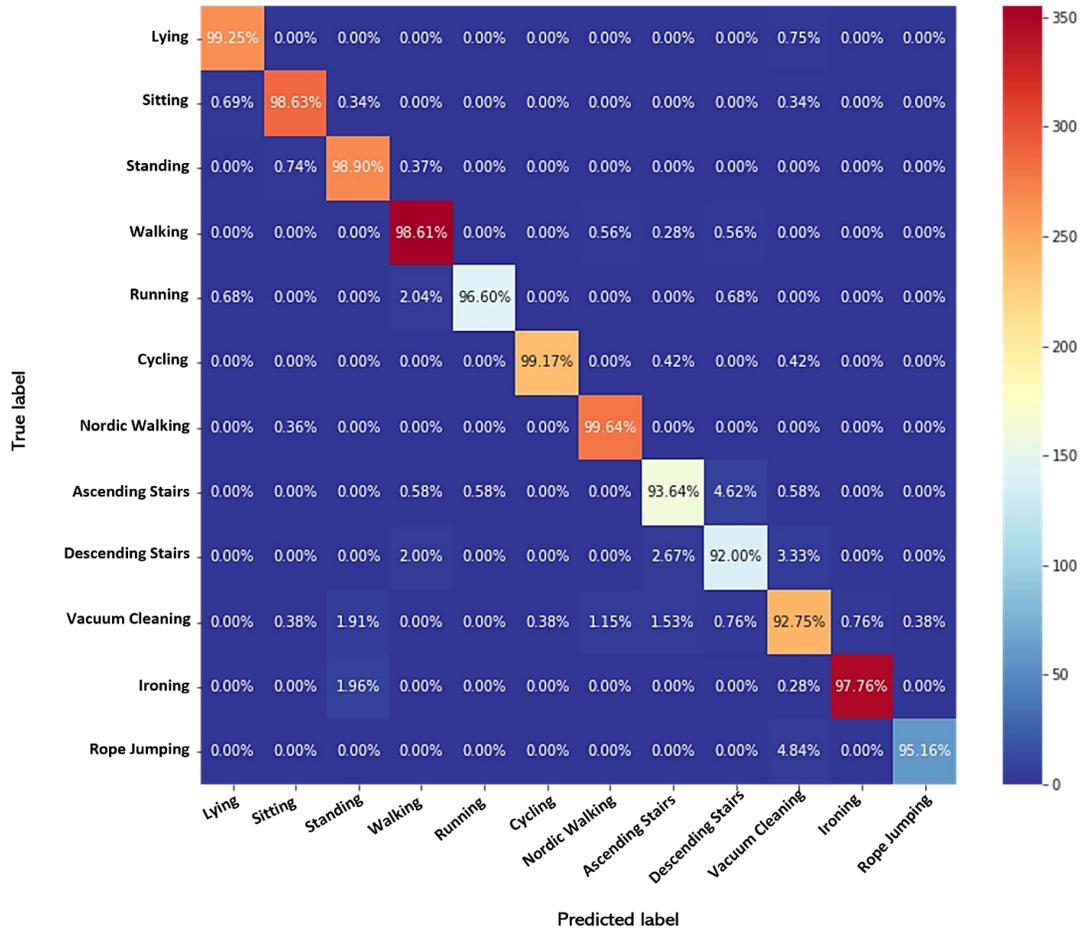

Figure 11: Confusion Matrix of 171 window size (ConvLSTM-WSense on PAMAP2)

As shown in the confusion matrix in Figure 11, lying activity had 99.25% of its samples classified correctly, while 0.75% were misclassified as vacuum cleaning. On sitting activity, 98.63% of the samples were correctly classified, with 0.34% misclassified as standing and another 0.34% as vacuum cleaning. Standing activity had 98.90% correctly classified samples, with 0.74% misclassified as sitting and 0.37% misclassified as walking. Likewise, walking activity had 98.61% of its samples correctly classified; 0.56% were misclassified as Nordic walking and descending stairs, while 0.28% were misclassified as ascending stairs. Similarly, running activity had 96.60% of its samples classified correctly, with 2.04% misclassified as walking and 0.68% misclassified as lying and descending stairs. Ascending stairs had 93.64% of its samples classified correctly, with 4.62% misclassified as descending stairs and 0.58% misclassified as walking, running, and vacuum cleaning each. Descending stairs had 92.00% correctly classified samples, with 2.00% misclassified as walking, 2.67% misclassified as ascending stairs, and 3.33% misclassified as vacuum cleaning. Ironing activity had 97.76% correctly classified samples, with 1.96% misclassified as standing and 0.285 misclassified as vacuum cleaning. Rope jumping had 95.16% correctly classified samples, while 4.84% were misclassified as vacuum cleaning.

## 4.5 Results on WISDM

Experiments were conducted on the WISDM dataset with varying sliding window sizes ranging between 80 to 360. The number of samples returned for each segmentation size is presented in Table 8.

Table 8: Training and testing sample per selected window size on WISDM

|       | Window Size | Training Samples | Testing Samples |
|-------|-------------|------------------|-----------------|
|       | 80          | 6887             | 1717            |
|       | 120         | 4577             | 1145            |
|       | 160         | 3432             | 859             |
| **WISDM** | 200     | 2746             | 687             |
|       | 240         | 2288             | 572             |
|       | 280         | 1960             | 491             |
|       | 320         | 1716             | 429             |
|       | 360         | 1524             | 382             |

### 4.5.1 CNN-WSense on WISDM

The model training and loss of the CNN-WSense on the best-performing window size are presented in Figure 12(a) and (b).

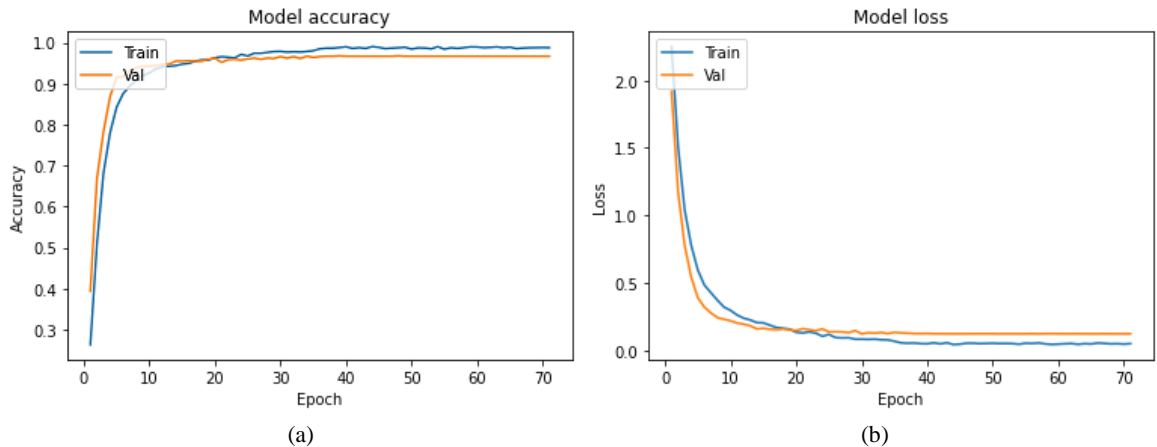

(a)             (b)
Figure 12: CNN-WSense on WISDM (a) Model training (b) Model loss

As stated earlier, ten (10) experiments were performed using each sliding window on the CNN-WSense model, the CNN-SE model, and the baseline CNN model. The average and highest accuracy of the three models is compared and presented in Table 9.

Table 9: CNN model comparison of average and highest accuracy on WISDM

| Window Size | Baseline CNN | | CNN-SE | | CNN-WSense | |
|---|---|---|---|---|---|---|
| | Average Accuracy(%) | Highest Accuracy(%) | Average Accuracy(%) | Highest Accuracy(%) | Average Accuracy(%) | Highest Accuracy(%) |
| 80 | 96.28 | 96.33 | 96.36 | 96.50 | **96.37** | **96.56** |
| 120 | 96.74 | 96.85 | 96.95 | 96.94 | **97.09** | **97.12** |
| 160 | 94.59 | 94.64 | 95.71 | 95.92 | **96.26** | **96.27** |
| 200 | 95.30 | 95.48 | 95.35 | 95.63 | **95.63** | **96.63** |
| 240 | 96.16 | 96.32 | 95.74 | 95.80 | **96.87** | **97.02** |
| 280 | 92.38 | 93.07 | 93.62 | 93.68 | **95.62** | **95.72** |
| 320 | 93.49 | 93.93 | 94.40 | 94.40 | **95.08** | **95.33** |
| 360 | 90.85 | 91.09 | 92.21 | 92.40 | **95.20** | **95.28** |
| 95% Confidence Interval | 94.47 ± 1.35 | | 95.04 ±0.902 | | **96.38 ±0.436** | |

The results in Table 9 showed that after ten (10) experiments on each sliding window, the CNN-WSense model had an average accuracy of 96.37% on window size 80, 97.10% on window size 120, 96.26% on 160, 95.63% on 200, 96.87% on 240, 95.62% on 280, 95.08% on 320, and 95.20% on 360 window size. These accuracies outperformed the performance recorded on the baseline model with a difference of 0.09%, 0.35%, 1.67%, 0.33%, 0.71%, 3.24%, 1.59%, and 4.35%. They outperformed the CNN-SE model with a difference of 0.01%, 0.14%, 0.55%, 0.28%, 1.13%, 2.00%, 0.68%, and 2.99% on window sizes 80, 120, 160, 200, 240, 280, 320 and 360 respectively. The 95% confidence interval of the CNN-WSense had 96.38 ±0.436, which shows a minimal difference of 0.436 in the accuracy achieved across all the window sizes, which outperformed the baseline recorded 94.47375 ± 1.35, and the CNN-SE, which had 95.04 ±0.902 confidence interval. The model size comparison and average accuracy of the CNN-WSense model, CNN-SE model, and Baseline CNN model on WISDM are presented in Table 10.

Table 10: Model size comparison on WISDM (CNN Models)

| Model Window Size | 80 | 120 | 160 | 200 | 240 | 280 | 320 | 360 | Accuracy |
|---|---|---|---|---|---|---|---|---|---|
| CNN | 0.727M | 1.055M | 1.383M | 1.710M | 2.038M | 2.366M | 2.694M | 3.021M | 94.47 ± 1.35 |
| CNN-SE | 0.732M | 1.059M | 1.387M | 1.715M | 2.042M | 2.370M | 2.698M | 3.025M | 95.04 ±0.902 |
| **CNN-WSense** | **0.236M** | **0.236M** | **0.236M** | **0.236M** | **0.236M** | **0.236M** | **0.236M** | **0.236M** | **96.38 ±0.436** |

From the results in Table 10, it can be noted that the parameters of the baseline CNN model increased as the window size increased, and it ranged between 727,942 to 3,021,702, directly affecting the training time and deployment on end devices. Similarly, the CNN-SE model had parameters ranging from 732,038 on window size 80 to 3,025,798 parameters on window size 360. However, with the CNN-WSense, a minimal parameter value of 236,678 was recorded across all sliding window sizes on the WISDM dataset. This shows that the model size was significantly reduced and remained uniform regardless of the size of the sliding window segmentation. The classification report on all experimented window sizes on the WISDM dataset by the CNN-WSense model is presented in Figure 13.

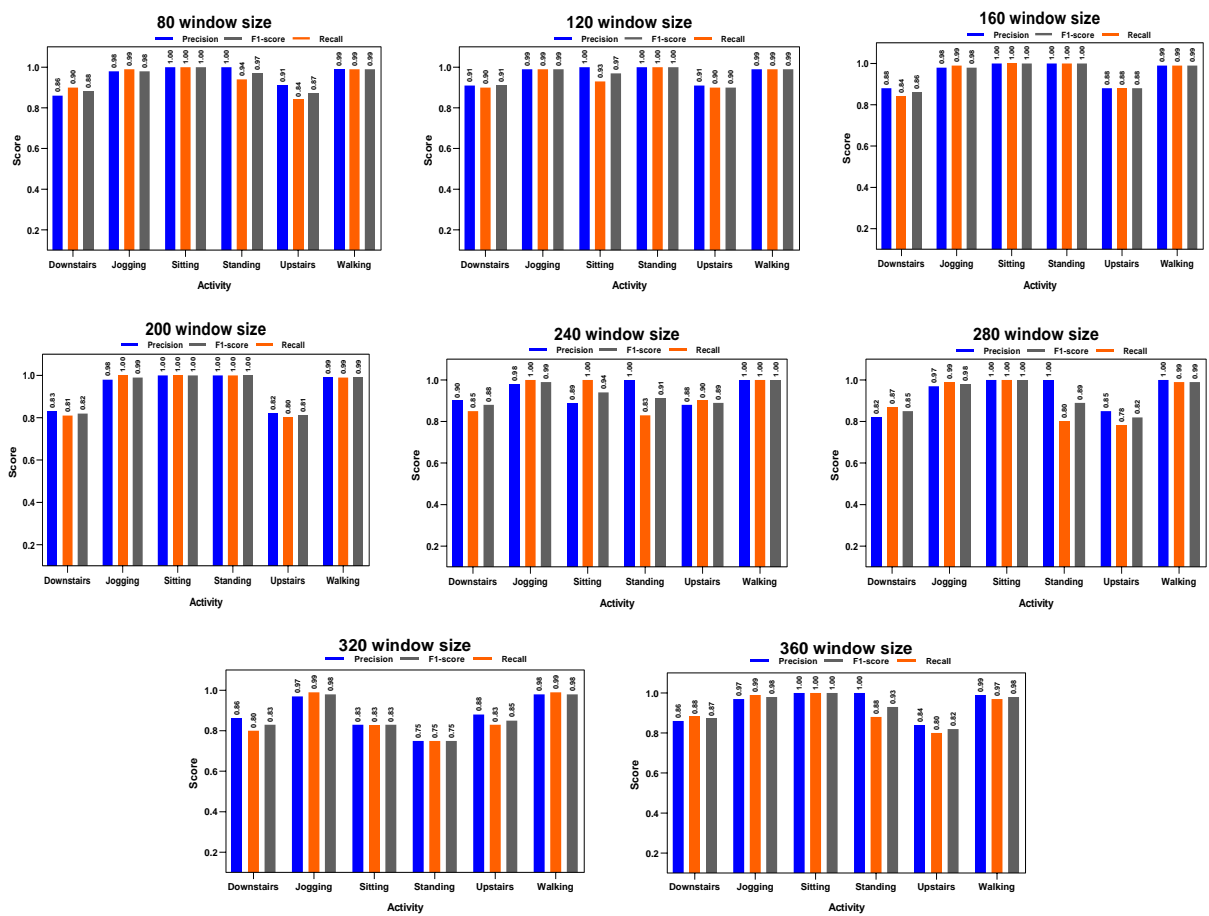

Figure 13: Classification Report of CNN-WSense across experimented window sizes on the WISDM Dataset

As shown in Table 9 and Table 10, the size of the baseline activity recognition model increased as the sliding window size increased. On the baseline CNN model on WISDM, sliding window size 80 had 727,942 parameters with an average accuracy of 96.28%, while CNN-SE had 732,038 parameters with an average recognition accuracy of 96.36%. Window size 120 had 1,055,622 parameters with an average accuracy of 96.74% on the baseline CNN, while CNN-SE had 1,059,718 with an average accuracy of 96.95%. Sliding window 160 had a parameter value of 1,383,302 with an average accuracy of 94.59% on baseline CNN, while CNN-SE had an accuracy of 95.71 with 1,387,398 parameters. Experiments on sliding window size 200 returned a model size of 1,710,982 parameters, with an average recognition accuracy of 95.30%, while CNN-SE had an average accuracy of 95.35 with 1,715,078 parameters. Similarly, sliding window size 240 had 2,038,662 parameters with an average recognition accuracy of 96.16%, while CNN-SE had a 95.74% average accuracy with 2,042,758 parameters.

On sliding window size 280, an average accuracy of 92.38% was achieved with 2,366,342 parameters, while CNN-SE had 2,370,438 parameters with 93.62% accuracy. Likewise, window size 320 achieved average recognition accuracy of 93.49% with 2,694,022 parameters, while the CNN-SE had a 94.40% average accuracy with 2,698,118 parameters. Finally, window size 360 with 3,021,702 parameters had an average activity recognition accuracy of 90.85%, while CNN-SE with 3,025,798 parameters had a 92.21% average accuracy. However, by using the CNN-WSense model for feature learning on WISDM, all the window sizes 80 – 360 had the same number of parameters at 236,678 (0.236million) parameters, and the average and highest activity recognition accuracy achieved using the CNN-WSense model on the WISDM dataset presented in Table 4 outperformed the baseline and CNN-SE models across all sliding window sizes, and this was achieved using a smaller and uniform number of parameters. The confusion matrix of the highest recognition accuracy achieved by the CNN-WSense model on WISDM across all window sizes is presented in Figure 14.

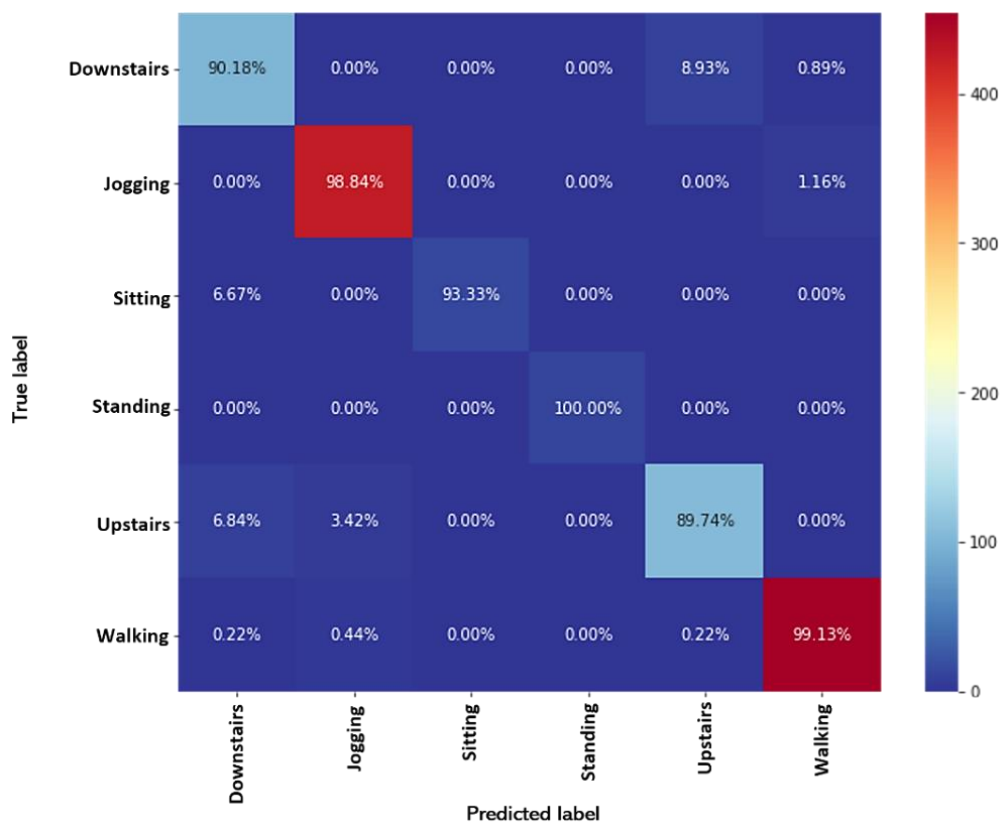

Figure 14: Confusion Matrix of 120 window size (CNN-WSense on WISDM)

As shown in the confusion matrix of 120 window size of CNN-WSense on WISDM presented in Figure 14, downstairs activity had 90.18% of its samples correctly classified, with 8.93% misclassified as upstairs and 0.89% as walking. Jogging activity had 98.84% of its samples correctly classified, with 1.16% misclassified as walking, while sitting had 93.33% of its samples correctly classified, with 6.67% misclassified as downstairs. Standing had all its samples correctly classified, while on upstairs activity, 89.74% of its samples were classified correctly, with 6.84% misclassified as downstairs and 3.42% as jogging, while walking had 99.13% correctly classified samples, with 0.44% misclassification as jogging, 0.22% as upstairs and 0.22% as downstairs. The comparison of the average accuracy achieved per window size using the CNN-WSense, CNN-SE, and baseline CNN on WISDM is presented in Figure 15.

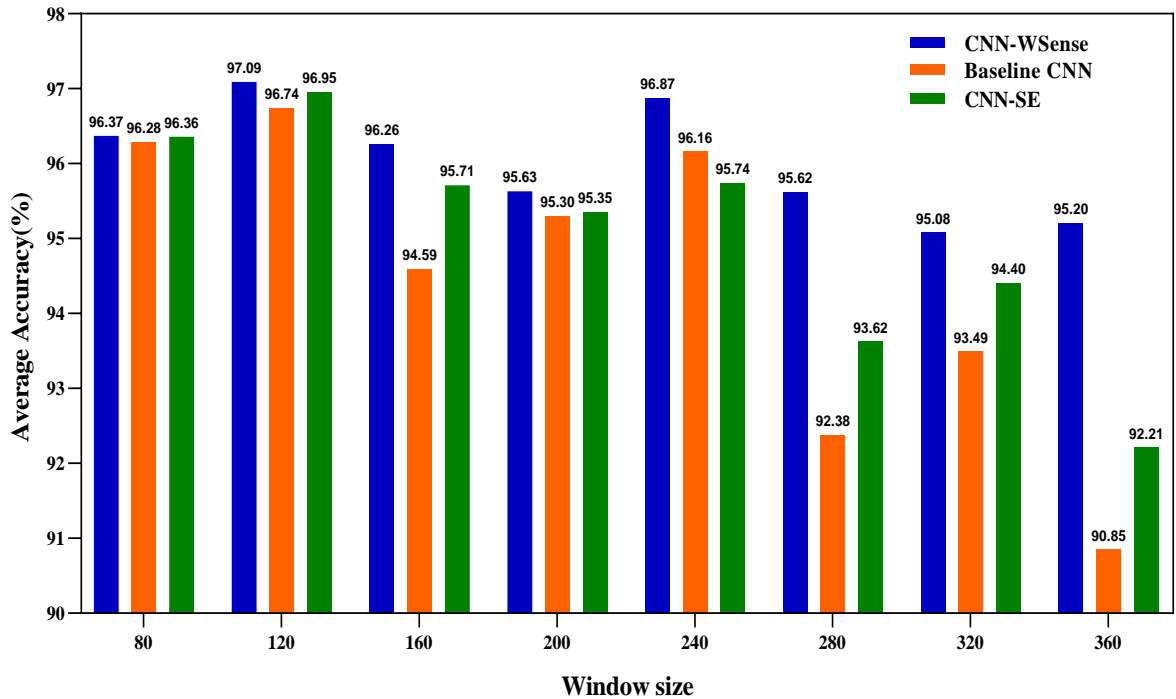

Figure 15: Average accuracy comparison of CNN models on WISDM window sizes

### 4.5.2 ConvLSTM-WSense on WISDM

Figures 16 (a) and (b) present the model training and loss on the best-performing window size.

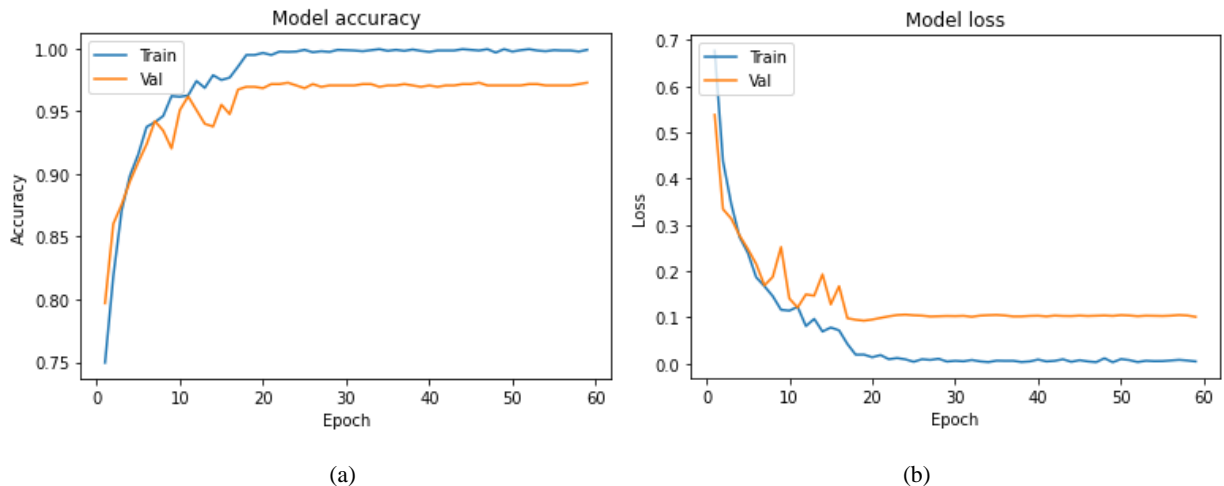

(a)                                          (b)

Figure 16: CNN-LSTM-WSense on WISDM (a) Model training (b) Model loss

On the ConvLSTM experiment on the WISDM dataset, ten experiments were run on the baseline model, excluding the WSense module, the ConvLSTM-SE model, which integrated squeeze and excitation block, and the ConvLSTM-WSense model. The average and highest accuracy obtained during experiments is shown in Table 11.

Table 11: ConvLSTM Comparison of average and highest accuracy on WISDM

|  | Baseline ConvLSTM | | ConvLSTM-SE | | ConvLSTM-WSense | |
| --- | --- | --- | --- | --- | --- | --- |
| Window Size | Average Accuracy(%) | Highest Accuracy(%) | Average Accuracy(%) | Highest Accuracy(%) | Average Accuracy(%) | Highest Accuracy(%) |
| 80 | 94.82 | 94.87 | 95.51 | 95.51 | **95.90** | **95.98** |
| 120 | 94.66 | 94.68 | 96.06 | 96.06 | **96.48** | **96.51** |
| 160 | 96.24 | 96.24 | 96.36 | 96.39 | **96.53** | **96.62** |
| 200 | 95.19 | 95.19 | 95.48 | 95.48 | **96.36** | **96.36** |
| 240 | 95.15 | 95.15 | 95.95 | 95.97 | **96.45** | **95.50** |
| 280 | 94.95 | 95.11 | 95.51 | 95.51 | **96.42** | **96.54** |
| 320 | 94.63 | 94.63 | 95.54 | 95.57 | **96.96** | **96.96** |
| 360 | 91.90 | 92.14 | 93.26 | 93.45 | **94.24** | **94.24** |
| 95% Confidence Interval | 94.69± 0.720 | | 95.45 ±0.550 | | **96.17 ±0.483** | |

The average accuracy of ten experiments per window size presented in Table 11 shows that the baseline ConvLSTM feature learning pipeline, which excluded the WSense module, recorded low accuracy compared to the feature learning pipeline of the ConvLSTM-WSense. Also, the ConvLSTM-SE model was outperformed by the ConvLSTM-WSense model. The comparison of the size of the activity recognition model returned per sliding window size is presented in Table 12.

Table 12: Model size comparison on WISDM (ConvLSTM Models)

|  | 80 | 120 | 160 | 200 | 240 | 280 | 320 | 360 | Accuracy |
| --- | --- | --- | --- | --- | --- | --- | --- | --- | --- |
| ConvLSTM | 0.504M | 0.635M | 0.832M | 0.963M | 1.160M | 1.291M | 1.487M | 1.618M | 94.69 ±0.720 |
| ConvLSTM-SE | 0.508M | 0.639M | 0.836M | 0.967M | 1.164M | 1.295M | 1.491M | 1.622M | 95.45 ±0.550 |
| **ConvLSTM-WSense** | **0.341M** | **0.341M** | **0.341M** | **0.341M** | **0.341M** | **0.341M** | **0.341M** | **0.341M** | **96.17 ±0.483** |

As shown in Table 12, the parameters of the baseline ConvLSTM and the ConvLSTM-SE models increased as the window size increased. The parameters of the baseline ConvLSTM ranged between 0.504 million to 1.618 million parameters on window size 80 to 360, which directly affects the training time and deployment on end devices. Similarly, the ConvLSTM-SE model had parameters ranging from 0.508 million on window size 80 to 1.622 million parameters on window size 360. However, the ConvLSTM-WSense recorded a minimal parameter value of 0.341 million across all sliding window sizes on the WISDM dataset. This shows that the model's size was significantly reduced and remained uniform regardless of the size of the sliding window segmentation. The classification report on all experimented window sizes on the WISDM dataset by the ConvLSTM-WSense model is presented in Figure 17.

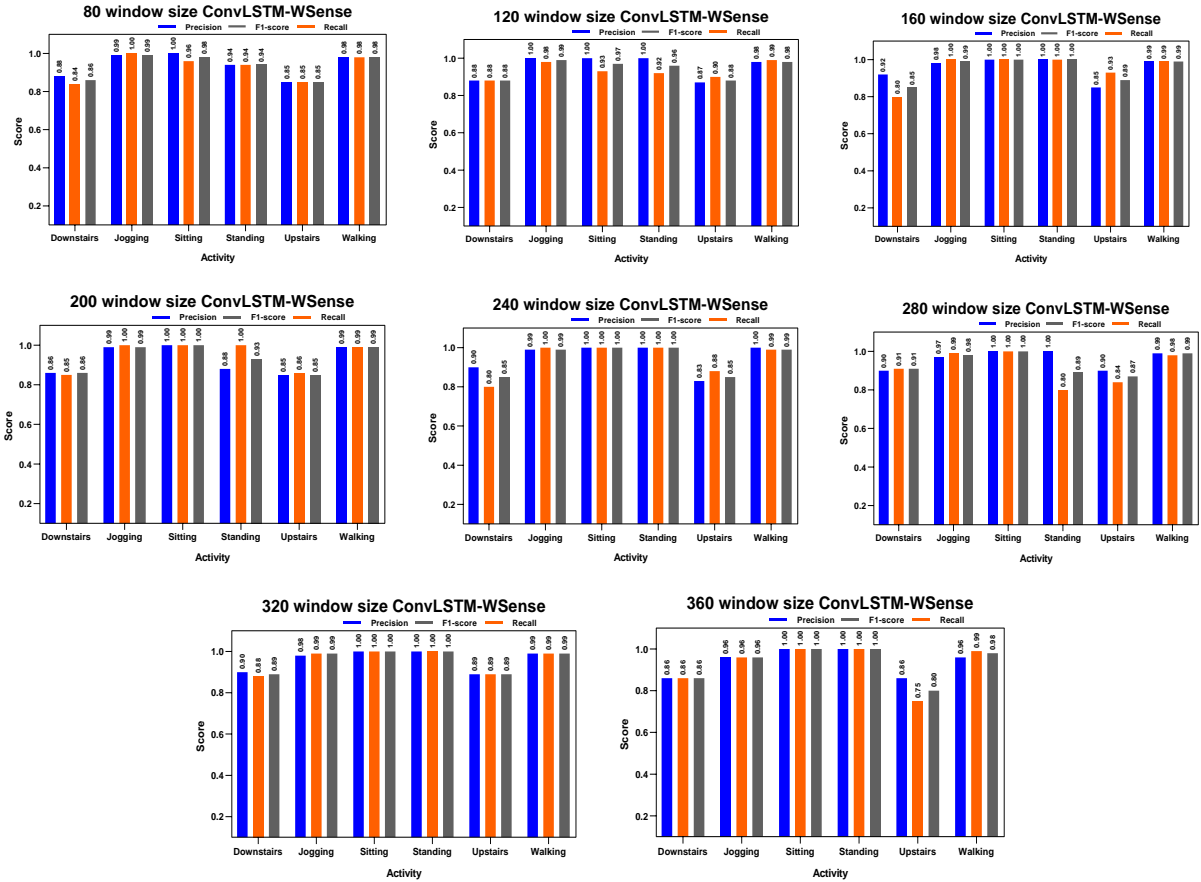

Figure 17: Classification Report of ConvLSTM-WSense across experimented window sizes on WISDM

As shown in Table 11 and Table 12, 80 sliding window sizes on the baseline ConvLSTM model returned an activity recognition model with 504,678 parameters with an average accuracy of 94.82% after ten experiments. While the ConvLSTM-SE model had 95.51% average accuracy with 508,774 model parameters. On the 120-sliding window size, the baseline ConvLSTM had a model with a size of 635,750 and average recognition accuracy of 94.66%, while ConvLSTM-SE had 639,846 parameters with 96.06% average accuracy. Sliding window size 160 returned an activity recognition model with 832,358 parameters and achieved average recognition accuracy of 96.24%, while ConvLSTM-SE had 96.36% with 836,454 model parameters. Using 200 sliding window size, the ConvLSTM model had a parameter value of 963,430, and an average activity recognition accuracy of 95.19%, while ConvLSTM-SE had 95.48% with 967,526 model parameters.

On Window size 240, the ConvLSTM model had 1,160,038 and an average recognition accuracy of 95.15, while ConvLSTM-SE had 95.95% average accuracy with 1,164,134 parameters. Similarly, on sliding window 280, the ConvLSTM model had 1,291,110 parameters with an average accuracy of 94.95%, while ConvLSTM-SE had a 95.51% average accuracy with 1,295,206 model parameters. On sliding window size 320, ConvLSTM had 1,487,718 model parameters, and an average accuracy of 94.63%, while the ConvLSTM-SE had 1,491,814 model parameters with 95.54% average accuracy. Finally, the baseline ConvLSTM model on 360 sliding window size had 1,618,790 parameters, with an average recognition accuracy of 91.90%, while ConvLSTM-SE had 1,622,886 million parameters with 93.26% average accuracy. However, the results of the experiments when the proposed WSense was plugged into the ConvLSTM feature learning pipeline showed that a model size of 341,094 (0.341million) was returned across all sliding window sizes, and the average accuracy achieved on each of the sliding window sizes outperformed the baseline ConvLSTM and ConvLSTM-SE models. The confusion matrix of the best recognition accuracy recorded using the CNN-LSTM-WSense is presented in Figure 18.

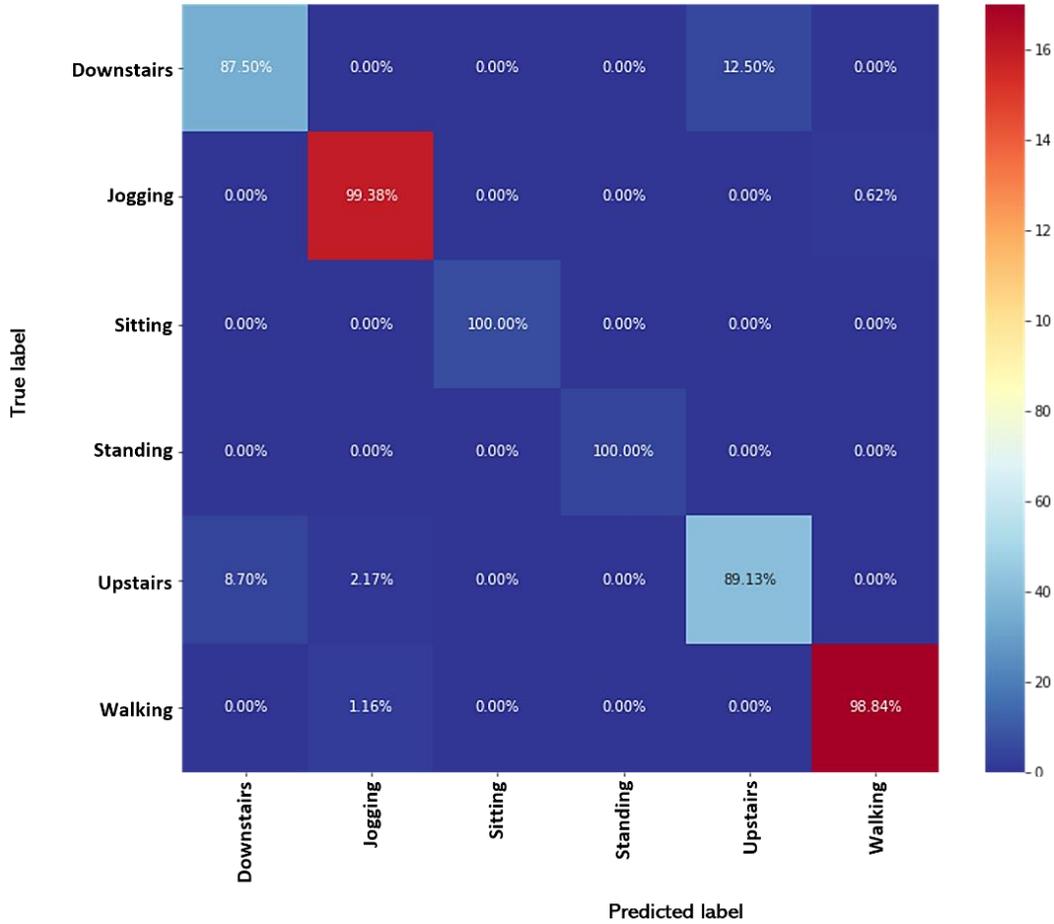

Figure 18: Confusion Matrix of 320 window size (ConvLSTM-WSense on WISDM)

The confusion matrix on window size 320, which achieved the highest accuracy using ConvLSTM-WSense, is presented in Figure 18. As shown, downstairs activity had 87.50% of its sample correctly classified, with 12.50% of the samples misclassified as upstairs. On jogging activity, 99.38% of the samples were correctly classified, with 0.62% misclassified as walking. Sitting and standing had 100% of samples classified correctly, while upstairs had 89.13% of samples correctly classified, 8.70% misclassified as downstairs, and 2.17% misclassified as jogging. On walking activity, 98.84% of the samples were correctly classified, with 1.16% misclassified as jogging. The average accuracy achieved across all window sizes is presented in Figure 19.

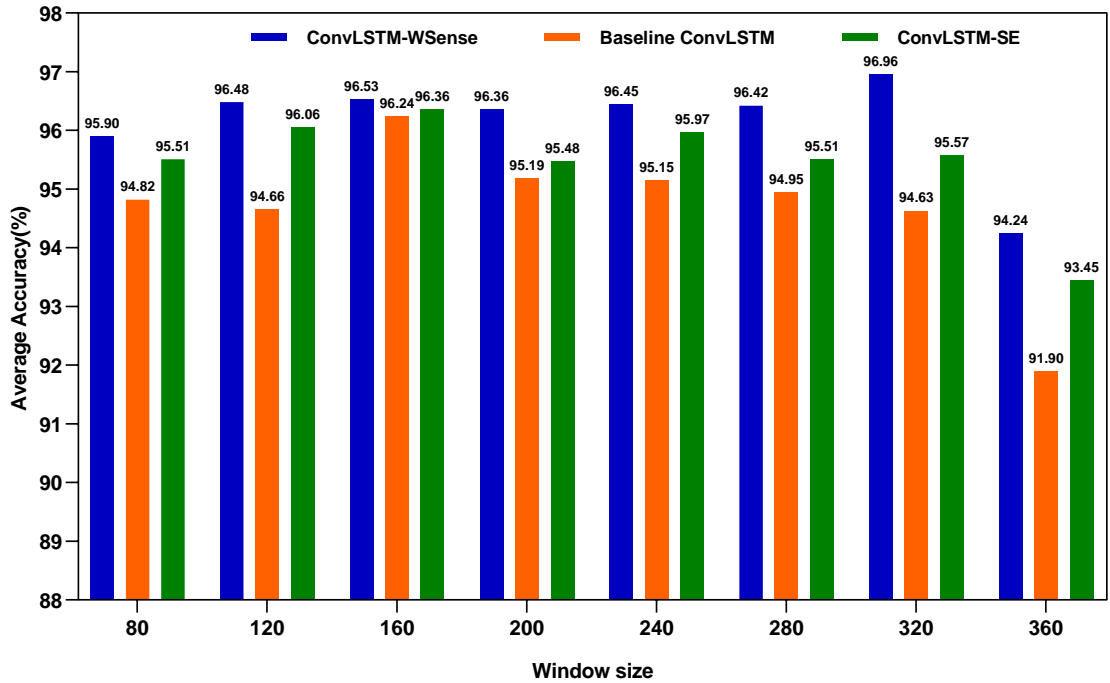

Figure 19: Average accuracy comparison of ConvLSTM models on WISDM window sizes

### 4.6 Comparison with the State-of-the-art Models

A comparison of the proposed WSense module with some state-of-the-art models in terms of their architecture, size, and results is presented in Table 13.

Table 13: Comparison with state-of-the-art activity recognition models

| Author | Method | Accuracy | Parameters |
|---|---|---|---|
| Teng *et al.* [49] | Layer-Wise Convolutional Neural Networks | PAMAP 92.97% | 2.60 million |
| Zhang *et al.* [34] | Multi-head feature learning pipeline with Squeeze and excitation block | WISDM 96.40% | 2.77 million |
| Gil-Martín *et al.* [50] | Sub-Window Convolutional neural network | PAMAP2 97.22% | 3.70 million |
| Gao *et al.* (2021) [21] | A combination of channel attention based on squeeze and excitation and temporal attention mechanisms to enhance discriminative feature extraction. | PAMAP2 93.16% | 3.51 million |
| Dua *et al.* (2021) [39] | A multiple-input CNN and Gated Recurrent Unit model for automatic feature extraction | PAMAP2 95.27% WISDM 97.21% | - - |

| Challa et al. [38] | Multiple branch CNN and Bi-LSTM model to extract features from activity signals. | PAMAP2 94.29% WISDM 96.05% | 0.647 million 0.622 million |
|---|---|---|---|
| Han et al. [45] | Different two-stream CNN architectures, which encode contextual information of activity signals from different receptive field sizes to generate more discriminative activity features | PAMAP2 92.97% | 1.37 million |
| Proposed WSense | CNN-WSense: | PAMAP2: 97.38% WISDM: 97.12% | 0.242 million 0.236 million |
| | ConvLSTM-WSense: | PAMAP2: 97.41% WISDM: 96.96% | 0.344 million 0.341 million |

As shown in Table 13, Gao et al. [21] proposed a dual attention model to improve feature learning in wearable sensor-based activity recognition and achieved a recognition accuracy of 93.16% on the PAMAP2 dataset, with 3.51M parameters. In Dua et al. [39], a CNN and GRU model with multiple inputs was proposed for improved feature learning from activity signals and achieved recognition accuracy of 95.24% on PAMAP2 and 97.21% on the WISDM dataset. Even though the size of the model was not presented in the research, the stacking structure of the layers shows that the size of the model will be bulky, as a fully connected layer was connected to the concatenation of the three-feature learning pipeline after two GRU layers, with no mechanism to reduce the size. Also, in Challa et al. [38], another multiple input model was proposed with CNN and Bi-LSTM and achieved recognition accuracy of 94.29% and 96.05% with 0.647M and 0.622M parameters on PAMAP2 and WISDM datasets, respectively. Likewise, in Han et al. [45], a heterogenous CNN module was proposed to improve feature learning in activity recognition and achieved an accuracy of 92.97% on PAMAP2 with 1.37M parameters. Even though these models are state-of-the-art, the limitation synonymous with them is the recognition accuracy recorded and the bulky size, which is a constraint when deploying activity recognition models on embedded devices. Aside from this, the models do not address the issues of varying model sizes and differences in learned features caused by sliding window segmentation size. However, these constraints were addressed with the WSense module plugged to feature learning pipelines, as minimal model sizes were recorded. At the same time, the module can achieve a state-of-the-art recognition accuracy of 97.38% and 97.12% on PAMAP2 and WISDM, respectively, by the CNN-WSense model. Also, the ConvLSTM-WSense had model parameters of 0.344M and 3.41M on PAMAP2 and WISDM, with highest recognition accuracy of 97.41% and 96.96% recognition respectively. Likewise, the WSense module was able to address the issue of varying model sizes, caused by the choice of sliding window segmentation size.

## 5.0 Conclusion

In this paper, we have proposed a feature learning module termed WSense to learn improved and similar features while ignoring the size of sliding window segmentation. The module was developed using 1-Dimensional convolutional layers and global max pooling layers. Experiments on the WISDM dataset and PAMAP2 showed that the WSense module improves feature learning using a minimal and uniform model size in activity recognition, regardless of the size of sliding window segmentation. Therefore, making models plugged with the WSense module easily deployable on end devices. Also, the difference in the learned features across all sliding window sizes reduced, as the 95% Confidence Interval on PAMAP2 using the CNN-WSense model was at **96.85 ± 0.141**, compared to the 95.07 ± 0.945 achieved on the baseline CNN and 96.05 ±0.539 achieved using CNN with squeeze and excitation. Similarly, on the WISDM dataset, a 95% CI of **95.775 ± 0.436** was achieved using the CNN-WSense, compared to the 94.47 ± 1.35 achieved on the baseline CNN and the 95.04 ±0.902 achieved by the CNN-SE model. Also, results of hybrid feature learning experiments in ConvLSTM-WSense on WISDM showed **96.17 ± 0.483** while 94.69 ± 0.720 was achieved on the baseline ConvLSTM and 95.45 ±0.550 on ConvLSTM-SE. Finally, on PAMAP2, the ConvLSTM-WSense model had a 95% Confidence Interval of **96.75 ± 0.241**, compared to the 95.79 ± 0.413 of its baseline model and the 96.29 ±0.257 of the ConvLSTM-SE model. The 95% Confidence Interval of the feature learning pipelines plugged with the WSense module proposed in this work shows that similar improved features were learned regardless of sliding window sizes compared to the baseline and squeeze and excitation models. Also, the WSense module ensured that the model size remained uniform and minimal, regardless of the sliding window size. By doing this, our research has proposed a method to ensure improved features can be learnt from wearable sensor signals, while ensuring the choice of the sliding window does not affect the quality of the extracted features and the size of the activity recognition model. Furthermore, WSense can be incorporated into bulky activity recognition models to reduce the model size while ensuring improved features are learned. For future work, the module will be evaluated on datasets with transitions to determine the module's viability when using adaptive sliding windows, which consider the length of transitional activities. Also, the extensibility of the proposed WSense module will be tested on other tasks, such as image classification and biomedical image segmentation. The code is available at https://github.com/AOige/WSense.

## Appendix

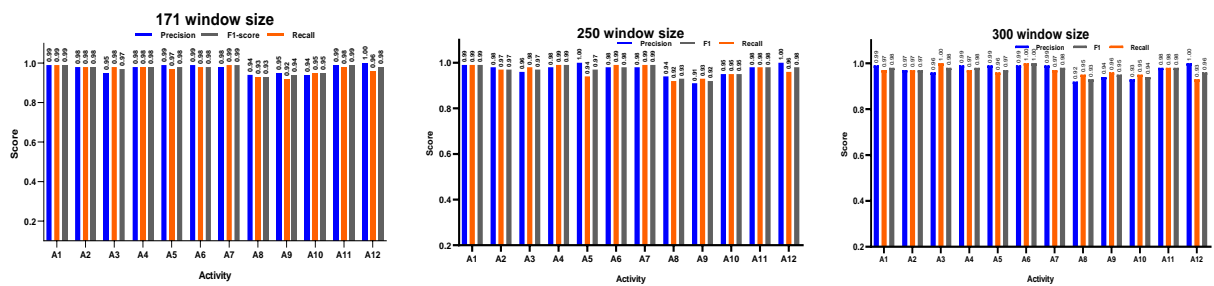

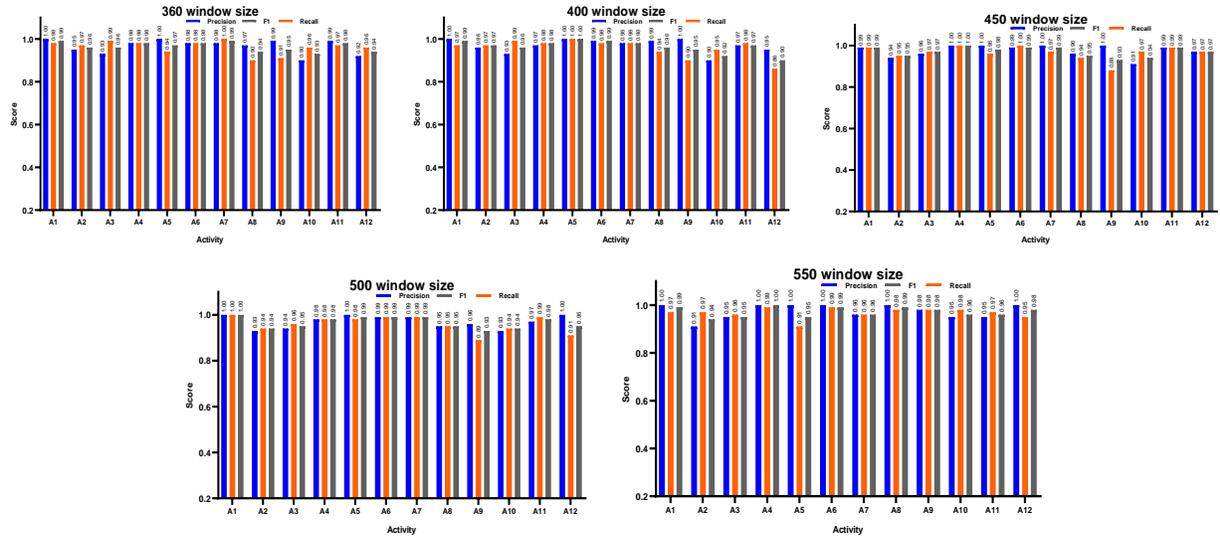

Appendix I: Classification Report of CNN-WSense across experimented window sizes on PAMAP2

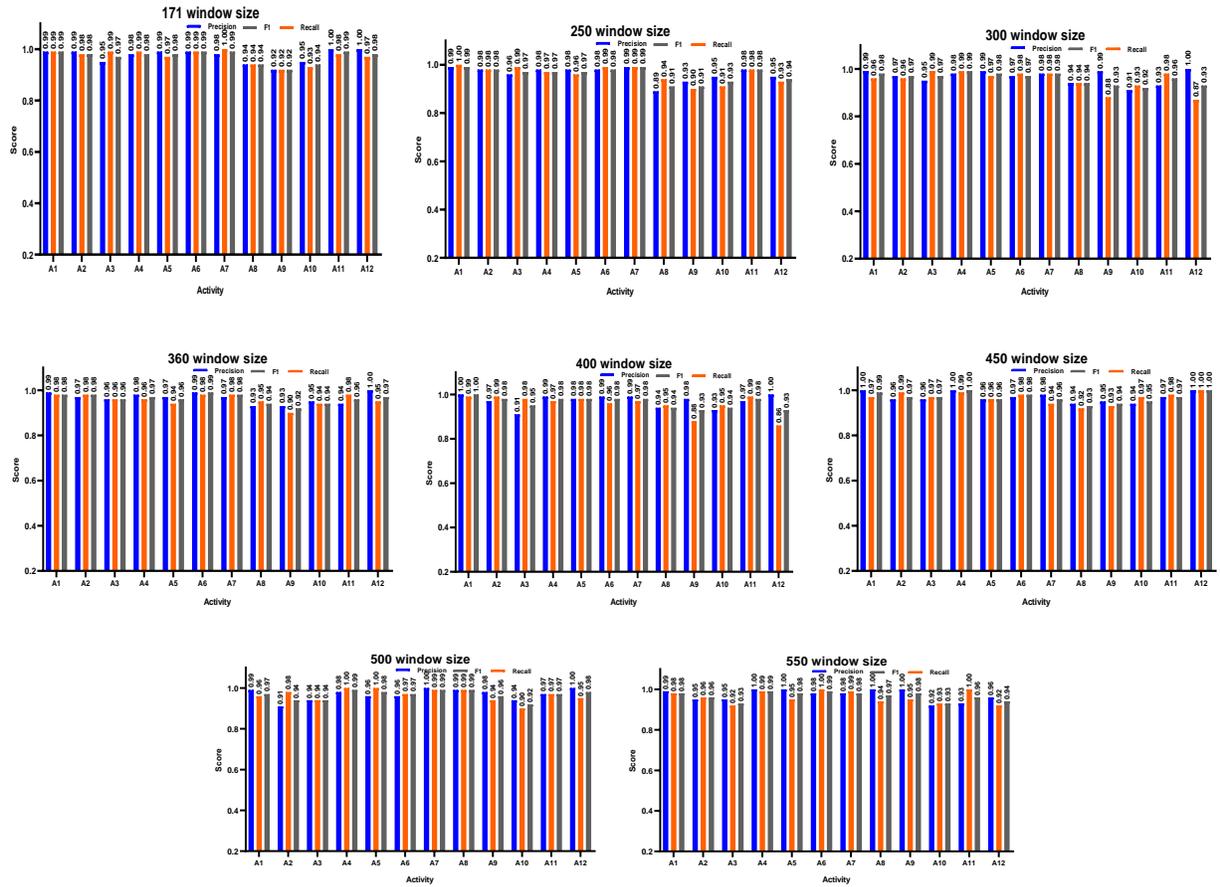

Appendix II: Classification Report of ConvLSTM-WSense across experimented window sizes on PAMAP2

# References

[1] J.D. Farah, N. Baddour, E.D. Lemaire, Design, development, and evaluation of a local sensor-based gait phase recognition system using a logistic model decision tree for orthosis-control, Journal of Neuroengineering and Rehabilitation. 16 (2019) 1–11.

[2] Y. Wang, S. Cang, H. Yu, A survey on wearable sensor modality centred human activity recognition in health care, Expert Systems with Applications. 137 (2019) 167–190. https://doi.org/10.1016/j.eswa.2019.04.057.

[3] O.D. Lara, M.A. Labrador, A Survey on Human Activity Recognition Using Wearable Sensors, in: IEEE Communications Surveys & Tutorials, 2013: pp. 1192–1209. https://doi.org/10.1029/GL002i002p00063.

[4] A.O. Adeola, O. Fapohunda, A.T. Jimoh, T.I. Toluwaloju, A.O. Ige, A.C. Ogunyele, Scientific applications and prospects of nanomaterials: A multidisciplinary review, African Journal of Biotechnology. 18 (2019) 946–961. https://doi.org/10.5897/ajb2019.16812.

[5] D. Ding, R.A. Cooper, P.F. Pasquina, L. Fici-Pasquina, Sensor technology for smart homes, Maturitas. 69 (2011) 131–136. https://doi.org/10.1016/j.maturitas.2011.03.016.

[6] M.H.M. Noor, A. Nazir, M.N.A. Wahab, J.O.Y. Ling, Detection of Freezing of Gait Using Unsupervised Convolutional Denoising Autoencoder, IEEE Access. 9 (2021) 115700–115709. https://doi.org/10.1109/ACCESS.2021.3104975.

[7] H. Gjoreski, M. Gams, M. Luštrek, Context-based fall detection and activity recognition using inertial and location sensors, Journal of Ambient Intelligence and Smart Environments. 6 (2014) 419–433. https://doi.org/10.3233/AIS-140268.

[8] S. Sani, N. Wiratunga, S. Massie, K. Cooper, kNN sampling for personalised human activity recognition, in: Lecture Notes in Computer Science (Including Subseries Lecture Notes in Artificial Intelligence and Lecture Notes in Bioinformatics), 2017: pp. 330–344. https://doi.org/10.1007/978-3-319-61030-6_23.

[9] M.T. Uddin, M.A. Uddiny, A guided random forest based feature selection approach for activity recognition, in: 2nd International Conference on Electrical Engineering and Information and Communication Technology, ICEEiCT 2015, Dhaka, 2015: pp. 1–6. https://doi.org/10.1109/ICEEICT.2015.7307376.

[10] K.G. Manosha Chathuramali, R. Rodrigo, Faster human activity recognition with SVM, International Conference on Advances in ICT for Emerging Regions, ICTer 2012 - Conference Proceedings. (2012) 197–203. https://doi.org/10.1109/ICTer.2012.6421415.

[11] L. Fan, Z. Wang, H. Wang, Human activity recognition model based on decision tree, in: Proceedings - 2013 International Conference on Advanced Cloud and Big Data, CBD 2013, Nanjing, 2013: pp. 64–68. https://doi.org/10.1109/CBD.2013.19.

[12] M.H.M. Noor, Z. Salcic, K.I.K. Wang, Adaptive sliding window segmentation for physical activity recognition using a single tri-axial accelerometer, Pervasive and Mobile Computing. 38 (2017) 41–59. https://doi.org/10.1016/j.pmcj.2016.09.009.

[13] O. Banos, J.M. Galvez, M. Damas, H. Pomares, I. Rojas, Window size impact in human activity recognition, Sensors (Switzerland). 14 (2014) 6474–6499. https://doi.org/10.3390/s140406474.

[14] K. Chen, D. Zhang, L. Yao, B. Guo, Z. Yu, Y. Liu, Deep learning for sensor-based human activity recognition: Overview, challenges, and opportunities, ACM Computing Surveys. 54 (2021) 1–40. https://doi.org/10.1145/3447744.

[15] M.H. Mohd Noor, S.Y. Tan, M.N. Ab Wahab, Deep Temporal Conv-LSTM for Activity Recognition, Neural Processing Letters. (2022). https://doi.org/10.1007/s11063-022-10799-5.

[16] B. Fida, I. Bernabucci, D. Bibbo, S. Conforto, M. Schmid, Varying behavior of different window sizes on the classification of static and dynamic physical activities from a single accelerometer, Medical Engineering and Physics. 37 (2015) 705–711. https://doi.org/10.1016/j.medengphy.2015.04.005.

[17] A. Bulling, U. Blanke, B. Schiele, A tutorial on human activity recognition using body-worn inertial sensors, ACM Computing Surveys. 46 (2014). https://doi.org/10.1145/2499621.
[18] A.O. Ige, M.H. Mohd Noor, A lightweight deep learning with feature weighting for activity recognition, Computational Intelligence. (2022).
[19] J. Hu, L. Shen, G. Sun, Squeeze-and-Excitation Networks, Proceedings of the IEEE Computer Society Conference on Computer Vision and Pattern Recognition. (2018) 7132–7141. https://doi.org/10.1109/CVPR.2018.00745.
[20] H. Ma, W. Li, X. Zhang, S. Gao, S. Lu, Attnsense: Multi-level attention mechanism for multimodal human activity recognition, IJCAI International Joint Conference on Artificial Intelligence. 2019-Augus (2019) 3109–3115. https://doi.org/10.24963/ijcai.2019/431.
[21] W. Gao, L. Zhang, Q. Teng, J. He, H. Wu, DanHAR: Dual Attention Network for multimodal human activity recognition using wearable sensors, Applied Soft Computing. 111 (2021) 107728. https://doi.org/10.1016/j.asoc.2021.107728.
[22] M. Abdel-Basset, H. Hawash, R.K. Chakrabortty, M. Ryan, M. Elhoseny, H. Song, ST-DeepHAR: Deep Learning Model for Human Activity Recognition in IoHT Applications, IEEE Internet of Things Journal. 8 (2021) 4969–4979. https://doi.org/10.1109/JIOT.2020.3033430.
[23] A.O. Ige, M.H. Mohd Noor, A survey on unsupervised learning for wearable sensor-based activity recognition, Applied Soft Computing. (2022) 109363. https://doi.org/10.1016/j.asoc.2022.109363.
[24] Ó.D. Lara, M.A. Labrador, A survey on human activity recognition using wearable sensors, IEEE Communications Surveys and Tutorials. 15 (2013) 1192–1209. https://doi.org/10.1109/SURV.2012.110112.00192.
[25] A. Ferrari, D. Micucci, M. Mobilio, P. Napoletano, Hand-crafted Features vs Residual Networks for Human Activities Recognition using Accelerometer, in: 2019 IEEE 23rd International Symposium on Consumer Technologies, ISCT 2019, 2019: pp. 153–156. https://doi.org/10.1109/ISCE.2019.8901021.
[26] M.H. Mohd Noor, Feature learning using convolutional denoising autoencoder for activity recognition, Neural Computing and Applications. 33 (2021) 10909–10922. https://doi.org/10.1007/s00521-020-05638-4.
[27] M. Yoshizawa, W. Takasaki, R. Ohmura, Parameter exploration for response time reduction in accelerometerbased activity recognition, UbiComp 2013 Adjunct - Adjunct Publication of the 2013 ACM Conference on Ubiquitous Computing. (2013) 653–664. https://doi.org/10.1145/2494091.2495986.
[28] M.S.H. Aung, S.B. Thies, L.P.J. Kenney, D. Howard, R.W. Selles, A.H. Findlow, J.Y. Goulermas, Automated detection of instantaneous gait events using time frequency analysis and manifold embedding, IEEE Transactions on Neural Systems and Rehabilitation Engineering. 21 (2013) 908–916. https://doi.org/10.1109/TNSRE.2013.2239313.
[29] S. Wan, L. Qi, X. Xu, C. Tong, Z. Gu, Deep Learning Models for Real-time Human Activity Recognition with Smartphones, Mobile Networks and Applications. 25 (2020) 743–755. https://doi.org/10.1007/s11036-019-01445-x.
[30] A. Dehghani, O. Sarbishei, T. Glatard, E. Shihab, A quantitative comparison of overlapping and non-overlapping sliding windows for human activity recognition using inertial sensors, Sensors (Switzerland). 19 (2019) 10–12. https://doi.org/10.3390/s19225026.
[31] M.M. Hassan, M.Z. Uddin, A. Mohamed, A. Almogren, A robust human activity recognition system using smartphone sensors and deep learning, Future Generation Computer Systems. 81 (2018) 307–313. https://doi.org/10.1016/j.future.2017.11.029.
[32] D. Anguita, A. Ghio, L. Oneto, X. Parra Perez, J.L. Reyes Ortiz, A public domain dataset for human activity recognition using smartphones, in: Proceedings of the 21th International European Symposium on Artificial Neural Networks, Computational Intelligence and Machine Learning, 2013: pp. 437–442.
[33] M.H.M. Noor, Z. Salcic, K.I.K. Wang, Dynamic sliding window method for physical activity recognition using a single tri-axial accelerometer, Proceedings of the 2015 10th IEEE Conference on

Industrial Electronics and Applications, ICIEA 2015. 38 (2015) 102–107. https://doi.org/10.1109/ICIEA.2015.7334092.

[34] H. Zhang, Z. Xiao, J. Wang, F. Li, E. Szczerbicki, A Novel IoT-Perceptive Human Activity Recognition (HAR) Approach Using Multihead Convolutional Attention, IEEE Internet of Things Journal. 7 (2020) 1072–1080. https://doi.org/10.1109/JIOT.2019.2949715.

[35] H. Khaled, O. Abu-Elnasr, S. Elmougy, A.S. Tolba, Intelligent system for human activity recognition in IoT environment, Complex and Intelligent Systems. (2021). https://doi.org/10.1007/s40747-021-00508-5.

[36] M. Kwabena Patrick, A. Felix Adekoya, A. Abra Mighty, B.Y. Edward, Capsule Networks – A survey, Journal of King Saud University - Computer and Information Sciences. 34 (2022) 1295–1310. https://doi.org/10.1016/j.jksuci.2019.09.014.

[37] S. Xiao, S. Wang, Z. Huang, Y. Wang, H. Jiang, Two-stream transformer network for sensor-based human activity recognition, Neurocomputing. 512 (2022) 253–268. https://doi.org/10.1016/j.neucom.2022.09.099.

[38] S.K. Challa, A. Kumar, V.B. Semwal, A multibranch CNN-BiLSTM model for human activity recognition using wearable sensor data, Visual Computer. (2021). https://doi.org/10.1007/s00371-021-02283-3.

[39] N. Dua, S.N. Singh, V.B. Semwal, Multi-input CNN-GRU based human activity recognition using wearable sensors, Computing. 103 (2021) 1461–1478. https://doi.org/10.1007/s00607-021-00928-8.

[40] L. Lu, C. Zhang, K. Cao, T. Deng, Q. Yang, A Multichannel CNN-GRU Model for Human Activity Recognition, IEEE Access. 10 (2022) 66797–66810. https://doi.org/10.1109/ACCESS.2022.3185112.

[41] R. Mutegeki, D.S. Han, A CNN-LSTM Approach to Human Activity Recognition, in: 2020 International Conference on Artificial Intelligence in Information and Communication (ICAIIC), 2020: pp. 362–366. https://doi.org/10.1109/ICAIIC48513.2020.9065078.

[42] M.A. Khatun, M.A. Yousuf, S. Ahmed, M.Z. Uddin, S.A. Alyami, S. Al-Ashhab, H.F. Akhdar, A. Khan, A. Azad, M.A. Moni, Deep CNN-LSTM With Self-Attention Model for Human Activity Recognition Using Wearable Sensor, IEEE Journal of Translational Engineering in Health and Medicine. 10 (2022) 1–16. https://doi.org/10.1109/JTEHM.2022.3177710.

[43] C. Xu, D. Chai, J. He, X. Zhang, S. Duan, InnoHAR: A deep neural network for complex human activity recognition, IEEE Access. 7 (2019) 9893–9902. https://doi.org/10.1109/ACCESS.2018.2890675.

[44] O. Nafea, W. Abdul, G. Muhammad, M. Alsulaiman, Sensor-based human activity recognition with spatio-temporal deep learning, Sensors. 21 (2021) 1–20. https://doi.org/10.3390/s21062141.

[45] C. Han, L. Zhang, Y. Tang, W. Huang, F. Min, J. He, Human activity recognition using wearable sensors by heterogeneous convolutional neural networks, Expert Systems with Applications. 198 (2022). https://doi.org/10.1016/j.eswa.2022.116764.

[46] Q. Xu, X. Wei, R. Bai, S. Li, Z. Meng, Integration of deep adaptation transfer learning and online sequential extreme learning machine for cross-person and cross-position activity recognition, Expert Systems With Applications. (2022) 118807. https://doi.org/10.1016/j.eswa.2022.118807.

[47] J.R. Kwapisz, G.M. Weiss, S.A. Moore, Activity recognition using cell phone accelerometers, ACM SIGKDD Explorations Newsletter. 12 (2011) 74–82. https://doi.org/10.1145/1964897.1964918.

[48] A. Reiss, D. Stricker, Introducing a new benchmarked dataset for activity monitoring, Proceedings - International Symposium on Wearable Computers, ISWC. (2012) 108–109. https://doi.org/10.1109/ISWC.2012.13.

[49] Q. Teng, K. Wang, L. Zhang, J. He, The Layer-Wise Training Convolutional Neural Networks Using Local Loss for Sensor-Based Human Activity Recognition, IEEE Sensors Journal. 20 (2020) 7265–7274. https://doi.org/10.1109/JSEN.2020.2978772.

[50] M. Gil-Martín, R. San-Segundo, F. Fernández-Martínez, J. Ferreiros-López, Time Analysis in Human Activity Recognition, Neural Processing Letters. 53 (2021) 4507–4525. https://doi.org/10.1007/s11063-021-10611-w.